\PassOptionsToPackage{table}{xcolor}
\documentclass[twocolumn]{bytedance_seed}

\usepackage[toc,page,header]{appendix}

\usepackage{minitoc}

\usepackage{graphicx}
\usepackage{amsmath}
\usepackage{amssymb}
\usepackage{multirow}
\usepackage{tabularx}
\usepackage{xcolor}
\usepackage{booktabs}
\usepackage{array}
\usepackage{stfloats}

\definecolor{oursrow}{HTML}{F2F2F2}
\newcolumntype{Y}{>{\raggedright\arraybackslash}X}

\title{Accelerating Unified Multimodal Models with Core-Expansion Routing and Unified Computation Scheduling}

\author[1,2,*]{Wengyi Zhan}
\author[2,*]{Chenqian Yan}
\author[2,\dagger,\ddagger]{Songwei Liu}
\author[1]{Mingbao Lin}
\author[1,\dagger]{Rongrong Ji}

\affiliation[1]{Xiamen University}
\affiliation[2]{ByteDance Seed}

\contribution[*]{Equal contribution}
\contribution[\dagger]{Corresponding authors}
\contribution[\ddagger]{Tech Lead}

\abstract{
Unified multimodal models jointly support understanding and generation, but incur substantial redundant computation across tokens, layers, and generation timesteps. Through token-importance probing, we identify an asymmetric core-expansion structure: understanding exhibits a stable importance component, while generation largely shares this component but requires progress-dependent corrections. We therefore propose \textbf{CE-Router}, which uses a task-shared core scorer and progress-conditioned generation expansions, optimized through generation decomposition and cross-task core alignment. At inference, CE-Router compacts token computation and supplies a learned routing signal to Unified Computation Scheduling, which coordinates layer skipping, FFN pruning, diffusion-head cache reuse, and denoising-step early exit. Experiments on two representative UMM architectures demonstrate consistent quality--efficiency improvements across both tasks, retaining 98.03\% of dense understanding performance with a 1.93$\times$ end-to-end inference speedup.
}

\date{\today}
\correspondence{
    Songwei Liu at \email{21831068@zju.edu.cn},
    Rongrong Ji at \email{rrji@xmu.edu.cn}
}

\begin{document}
\maketitle


\section{Introduction}
\label{sec:introduction}

Unified multimodal models (UMMs) increasingly support multimodal
understanding and visual generation within a single modeling
framework~\cite{deng2025bagel,xie2025showo,wang2024emu3,team2024chameleon}. Their implementations vary: some use a shared transformer for both tasks~\cite{xie2025showo,wang2024emu3}, while others couple them through mixture-of-experts or partially shared computation~\cite{shazeer2017moe,deng2025bagel,wu2025janus}.
Despite these architectural differences, understanding and generation are no longer isolated pipelines. They now execute through shared or tightly coupled backbone computation, creating new opportunities for reuse but also making inference efficiency inherently cross-task.

\begin{figure*}[!ht]
  \centering
  \includegraphics[width=\linewidth]{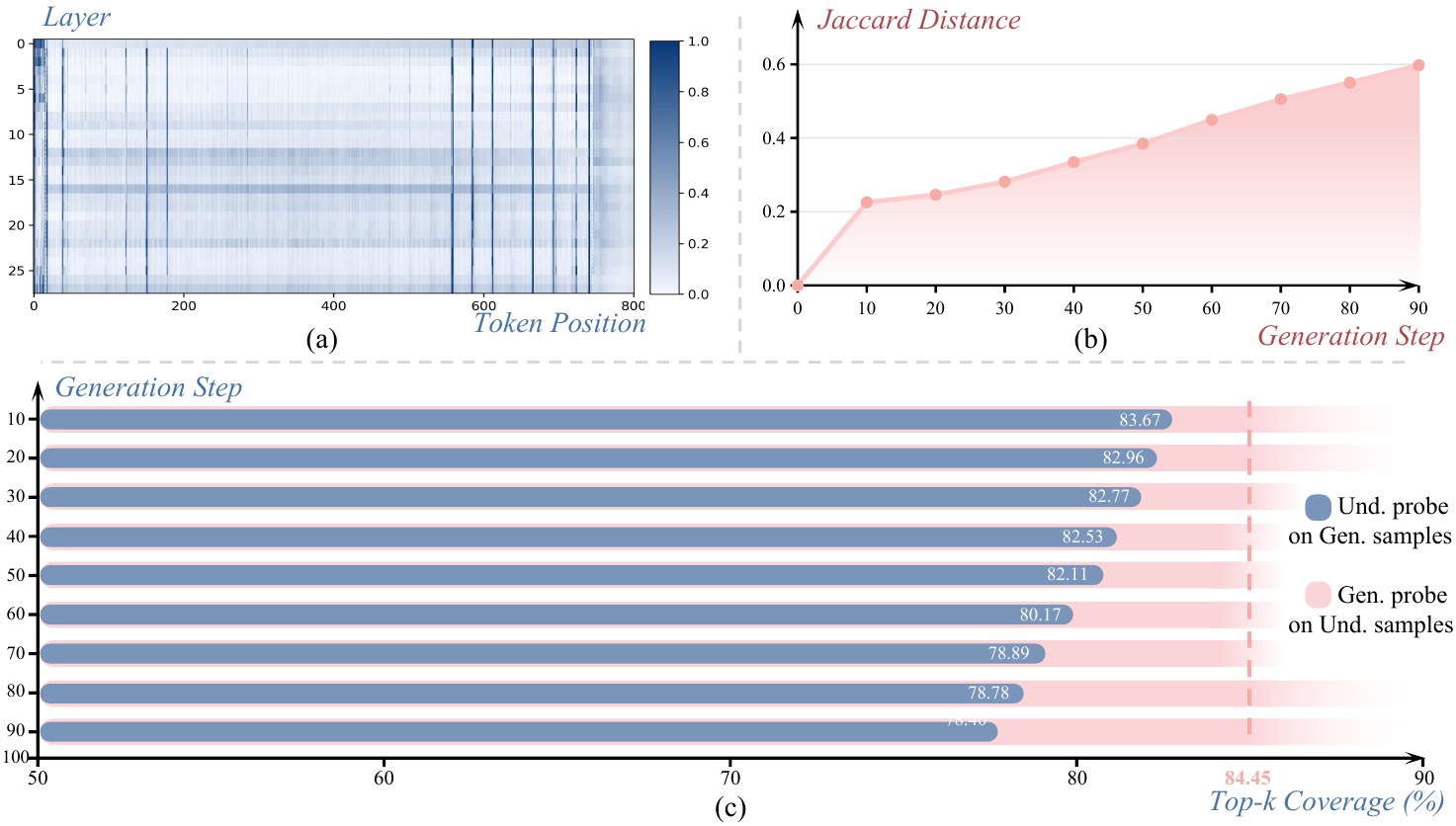}
  \caption{
    Token-importance analysis on Show-o2~\cite{xie2026showo2}.
    (a) Token-wise received attention across understanding layers.
    (b) Jaccard distance between the generation top-$k$ token set at each denoising timestep and the initial-timestep set. 
    (c) Directional transfer between task-specific probes, measured by
    top-$k$ coverage. Blue bars show the fraction of generation-probe
    selections recovered by the understanding probe at each timestep; the pink dashed line shows the reverse coverage.
    }
  \label{fig:intro1}
\end{figure*}

UMMs inherit redundant computation along three intertwined axes.
\textbf{(i) Context inflation.}  Visual inputs introduce long token sequences, although only a subset is strongly task-relevant. Redundant tokens enlarge attention and FFN computation and, for understanding, also produce a larger key--value cache that is repeatedly accessed during autoregressive decoding. \textbf{(ii) Depth-wise over-updating.}  Dense inference updates every token
at every layer even when many representations have stabilized or contribute little to the task output. \textbf{(iii) Trajectory re-computation.}  Visual generation repeatedly processes long sequences over many denoising steps, causing small per-step inefficiencies to accumulate into substantial latency.  Since these costs occur inside the same or coupled backbone, effective acceleration requires more than pruning each task separately: it requires determining which importance structure is reusable across tasks and which part changes with generation progress.

Existing acceleration methods only partially address this cross-task
structure.  Input-level approaches compress visual tokens before they enter the backbone, while trajectory-level approaches shorten generation through distillation or related mechanisms~\cite{wang2026unicompress,xu2025showoturbo,lu2025hyperbagel,li2026g2tr}. These methods reduce input or trajectory cost effectively, but operate before or around the shared backbone. They therefore do not characterize how redundant computation is organized once understanding and generation pass through shared or coupled internal layers.

Within-backbone methods~\cite{he2025understanding} move closer to this
problem through training-free shortcuts or learned routing.  Training-free frameworks such as FlashU~\cite{ke2026flashu} combine multiple acceleration operators, but their decisions rely on hand-crafted indicators and operator-specific operating points calibrated outside the task objective. Even when individual rules adapt across inputs or timesteps, they are not coordinated by a common learned notion of importance.  Learned routing methods~\cite{mao2025unimod} provide task-aligned decisions, but typically instantiate independent routers for understanding and generation.  This design captures task differences but duplicates scoring structure that may already be shared.  At the other extreme, a fully shared router is compact
but cannot represent progress-dependent changes unique to generation~\cite{mao2025unimod}.
Consequently, existing designs decide in advance whether routing should be shared or separated rather than deriving that boundary from the model's cross-task redundancy. This has motivated our central question in this paper: \emph{are understanding and generation redundancy patterns independent, identical, or related through a structured sharing boundary?}

\begin{figure}[t]
  \centering
  \includegraphics[width=\linewidth]{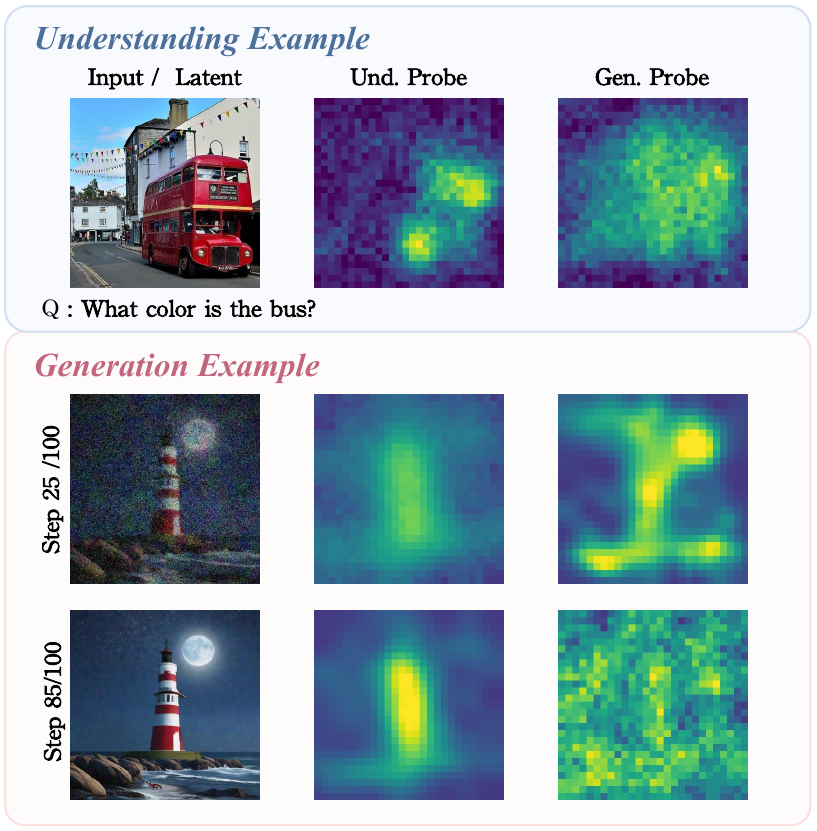}
  \caption{Qualitative visualization of asymmetric cross-task token importance. Both task-specific probes emphasize the task-relevant region in an understanding or generation example. Across two timesteps in the generation process, the understanding probe retains a stable semantic focus, whereas the generation probe varies with denoising progress. Note that brighter regions indicate higher importance.}
  \label{fig:visual1}
\end{figure}

To answer this question, we conduct three probes on a representative unified backbone. First, persistent received-attention patterns across layers suggest a stable component in understanding-side token importance (Figure\,\ref{fig:intro1}(a)). Second, the increasing Jaccard distance of a generation probe's top-$k$ set from its initial-timestep set shows that generation importance varies with denoising progress (Figure\,\ref{fig:intro1}(b)).  Third, directional transfer is asymmetric: the generation-trained probe recovers much of the understanding target set, whereas understanding-to-generation coverage declines over time (Figure\,\ref{fig:intro1}(c)).  Figure\,\ref{fig:visual1} provides a qualitative example.

These results show that the tasks are neither redundancy-identical nor independent. They share a transferable importance component, while generation requires progress-conditioned residual corrections. We call this common-plus-residual relation a \emph{core-expansion} redundancy structure.

Motivated by this structure, we propose the \textbf{Core-Expansion Router (CE-Router)}, a trainable token-importance module inserted into routed backbone layers.  Each router contains a core scorer shared by understanding and all generation segments, plus generation-only expansions selected by the current timestep.  The core is an input- and layer-dependent scoring function, not a fixed token subset; generation adds a segment-specific residual to its score.  This design preserves shared capacity while allocating additional parameters only to progress-dependent variation.

Because the intended functional specialization does not emerge naturally in a single training stage, we use two-stage optimization.  Stage I trains on generation across all denoising segments: every sample updates the core, while only its active expansion is updated. The core therefore captures recurring patterns and the expansions specialize to segment-specific deviations.  Stage II jointly trains both tasks; understanding updates only the core, whereas generation updates the core and active expansion. Both stages use the original task objective and dense-execution consistency.

At inference, CE-Router converts its logits into retained and dropped token sets, gathering only retained tokens for attention and FFN computation. Understanding routes during prefill and reuses the compressed key--value cache during decoding, while generation recomputes its retained set at every timestep.
Beyond token compaction, we introduce \emph{Unified Computation Scheduling} (UCS), which reuses CE-Router decisions across four acceleration operators: (1) layer skipping, (2) FFN pruning, (3) diffusion-head cache reuse, and (4) denoising-step early exit.  Retained tokens calibrate layer and FFN decisions, while drop-set stability guides cache reuse and generation termination. Thus, otherwise separate operators are conditioned on the same task-loss-aligned routing interface.

Our contributions are summarized as follows:
\begin{enumerate}
    \item We characterize token importance across model depth, denoising progress and modality tasks, revealing an asymmetric core–expansion redundancy structure.

    \item We introduce \textbf{CE-Router}, which combines a task-shared core with progress-conditioned generation expansions and is trained via a two-stage optimization strategy.

    \item We develop \textbf{Unified Computation Scheduling (UCS)} to coordinate four complementary acceleration operators through the common learned routing signal.
\end{enumerate}

Experiments on Show-o2 and BAGEL show that CE-Router provides
the strongest task performance among accelerated baselines, while UCS
further improves efficiency.  The full framework achieves up to a
1.93$\times$ understanding speedup while retaining 98.03\% of dense
performance.


\section{Related Work}

\subsection{Unified Multimodal Models (UMMs)}

Unified multimodal models place understanding and generation within a common interface or backbone.
Unified-IO 2~\cite{lu2024unifiedio2} supports heterogeneous modalities through one encoder--decoder, while SEED-LLaMA and SEED-X~\cite{ge2024seedllama,ge2024seedx} use discrete visual tokens for both comprehension and generation.
Chameleon~\cite{team2024chameleon} models interleaved text--image sequences with a single transformer; VILA-U~\cite{wu2025vila} and Emu3~\cite{wang2024emu3} formulate both tasks through unified next-token prediction.
Transfusion~\cite{zhou2025transfusion} combines language-modeling and diffusion objectives in one transformer, while Show-o~\cite{xie2025showo} couples autoregressive text modeling with discrete image diffusion.
BAGEL~\cite{deng2025bagel} further scales unified pretraining through partially shared computation.
Their degree of sharing nevertheless varies: Chameleon, Emu3, and Show-o strongly couple the two tasks, whereas Janus~\cite{wu2025janus} separates its visual encoders while retaining a unified autoregressive backbone.
Whereas these works primarily address architecture, pretraining, and task compatibility, we study how unification structures computational redundancy across tasks.

\subsection{Efficient Computation for UMMs}

Task-specific acceleration typically prunes visual tokens for understanding~\cite{chen2024fastv,shang2025llavaprumerge,yang2025visionzip} and uses caching, distillation, or fewer sampling steps for generation~\cite{ma2024deepcache,liu2025timestep,peng2025ertacache,xu2026motion}.
UMM-oriented methods extend these ideas at either the model boundary or within the backbone.
UniCompress~\cite{wang2026unicompress} and G$^2$TR~\cite{li2026g2tr} reduce input visual tokens, while Show-o Turbo~\cite{xu2025showoturbo} and Hyper-Bagel~\cite{lu2025hyperbagel} shorten generation trajectories.
Because they act before or outside shared backbone computation, these methods do not characterize which internal redundancy is shared across tasks.

Within-backbone approaches expose complementary limitations.
FlashU~\cite{ke2026flashu} combines pruning, layer skipping, and feature caching through training-free heuristics, while sparsity analyses identify exploitable component- and parameter-level activation patterns~\cite{he2025understanding}.
Such methods rely on separate indicators or thresholds rather than a common task-aligned signal.
Learned conditional computation addresses calibration: Mixture-of-Depths~\cite{raposo2024mod} routes tokens across layers, and UniMoD~\cite{mao2025unimod} introduces task-specific routers for UMMs.
However, independent routers duplicate task-shared structure instead of explicitly modeling it, and overly task-specific token reduction may weaken cross-task synergy~\cite{chen2026limits}.
CE-Router instead learns the sharing boundary itself and reuses that signal to coordinate conditional computation across both tasks.

\section{Methodology}

\begin{figure*}[!t]
  \centering
  \includegraphics[width=\linewidth]{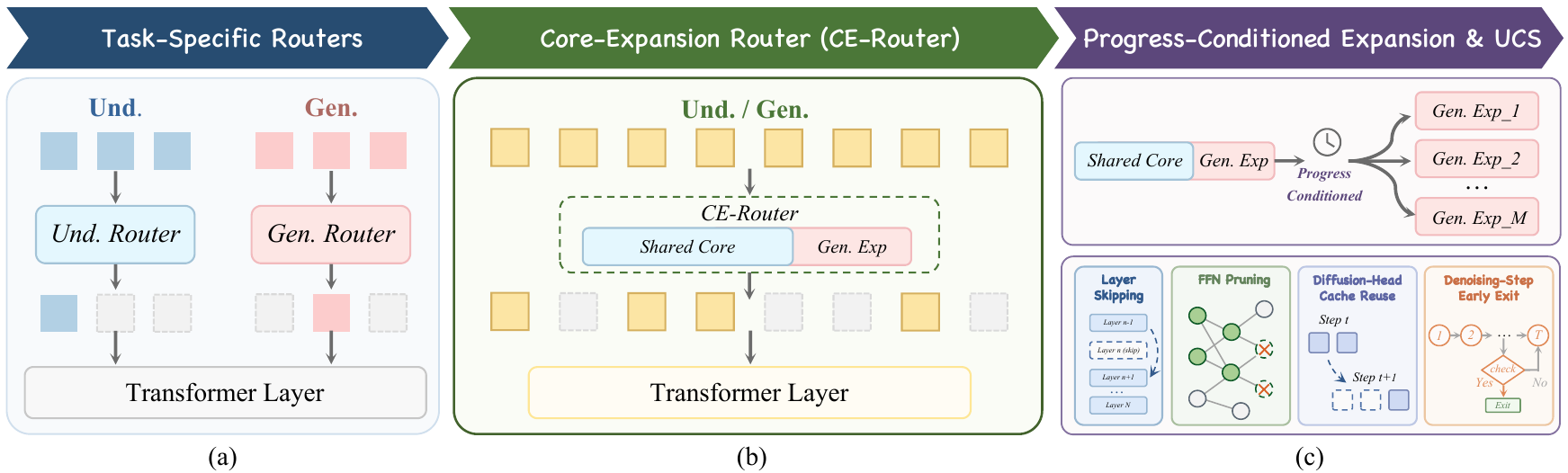}
  \caption{
    Pipeline of Core-Expansion Routing and Unified Computation Scheduling.
    (a) Conventional task-specific routers.
    (b) CE-Router with a task-shared core and generation expansions.
    (c) Progress-conditioned expansion selection and UCS operators.
  }
  \label{fig:pipeline}
\end{figure*}

We first probe token importance across model depth, denoising progress, and tasks,
revealing an asymmetric structure: understanding relies on a stable shared
component, while generation requires progress-dependent variation. CE-Router models this structure with a task-shared core and generation-specific expansions optimized through two-stage training.  At inference, its routing decisions guide token compaction and Unified Computation Scheduling across layers, FFN channels, caches, and denoising steps.
The overall architecture is illustrated in Figure\,\ref{fig:pipeline}.

\subsection{Probing Asymmetric Cross-Task Redundancy}
\label{sec:method_motivation}

\noindent\textbf{Depth-wise persistence in understanding.} We first examine 300 randomly sampled understanding examples using the dense, uncompressed model.  For input $x$, let $A_l(x)\in\mathbb{R}^{N\times N}$ be the head-averaged attention map at layer $l$, where $A_l(x)_{j,i}$ denotes the attention from query token $j$ to token $i$.  The attention received by token $i$ is
\begin{equation}
    r_{l,i}(x)
    =
    \frac{1}{N}
    \sum_{j=1}^{N}
    A_l(x)_{j,i}.
\end{equation}

Figure\,\ref{fig:intro1}(a) visualizes the sample-averaged scores.  The pronounced vertical bands persist across most of the backbone, indicating that some token positions repeatedly attract high attention.  Although received attention is only a proxy for routing importance, this persistence supports a stable component in understanding-side token scoring.

\noindent\textbf{Trajectory-wise variation in generation.}
Generation exhibits a different temporal pattern.  At each evaluated denoising timestep $t$, a generation-trained probe (see Appendix B for training details) selects the top-$k$ token set $S_G^t$. We quantify its departure from the initial selection $S_G^{t_0}$ using the Jaccard distance
\begin{equation}
    D_J(t)
    =
    1-
    \frac{
        |S^t_G\cap S^{t_0}_G|
    }{
        |S^t_G\cup S^{t_0}_G|
    },
\end{equation}
averaged over samples and layers.

As shown in Figure\,\ref{fig:intro1}(b), the distance increases as denoising advances: tokens emphasized near the beginning are progressively replaced by tokens relevant to later stages.  Consequently, a single generation-wide routing decision cannot represent the full trajectory.

\noindent\textbf{Directional cross-task transfer.}
The preceding analyses do not reveal whether either task-specific importance pattern transfers to the other. We therefore train architecture-matched probes for understanding and generation, and let $S_T(x)$ be the top-$k$ set selected on input $x$ by the probe trained for $T\in\{U,G\}$.  For generation input $x_G^t$, the understanding-to-generation coverage is:
\begin{equation}
\mathrm{Coverage}(U\rightarrow G;t)
=
\frac{
|S_U(x_G^t)\cap S_G(x_G^t)|
}{|S_G(x_G^t)|}
\times 100\%.
\end{equation}

We define $\mathrm{Coverage}(G\rightarrow U)$ analogously on understanding inputs, yielding one reference value because understanding has no denoising trajectory.  Averaged over samples and layers, Figure\,\ref{fig:intro1}(c) shows that the generation-trained probe recovers $84.45\%$ of the understanding target set, whereas $\mathrm{Coverage}(U\rightarrow G;t)$ decreases from $83.67\%$ to $78.40\%$ during denoising.  Thus, generation-oriented importance transfers more completely to understanding than the reverse at later generation stages.

Together, the probes support an asymmetric sharing structure rather than an exact containment relation.  A fully shared scorer can reuse the common signal but cannot model generation-side variation; independent task-specific scorers can model that variation but duplicate the transferable component. We therefore introduce Core-Expansion Scoring, comprising a task-shared core and lightweight, progress-conditioned residual scorers for generation.

\subsection{Core-Expansion Scoring}
\label{sec:method_scoring}

Let $H_l=[h_{l,1},h_{l,2},\dots,h_{l,N}]$, with $h_{l,i}\in\mathbb{R}^{d}$, denote the token representations entering a routed backbone layer $l$.  The router at this layer contains one \textit{core scorer} $c_l(\cdot;\theta_l^{c})$ and $M$ \textit{generation
expansion scorers} $e_{l,m}(\cdot;\theta_{l,m}^{e})$, all implemented as lightweight token-wise MLPs with scalar outputs.

The term \textit{shared core} refers specifically to cross-task parameter sharing.  At layer $l$, the same parameters $\theta_l^{c}$ are used by the understanding path and by every generation segment; there is no separate understanding core or segment-specific copy.  The scorers remain layer-specific, however.  Moreover, the core is a scoring function rather than a fixed set of ``core tokens,'' so its output remains input-dependent through $h_{l,i}$. In contrast, each expansion is generation-only and receives parameters tied to one denoising segment, establishing the intended common-plus-residual structure at every routed layer.

To model the remaining generation-side variation, we partition the discrete denoising timesteps into $M$ segments $\{\mathcal{T}_m\}_{m=1}^{M}$ and associate one expansion scorer with each segment.  Let $m(t)$ denote the segment containing timestep $t$.  Omitting parameters from the function arguments for readability, the routing logits are
\begin{equation}
\begin{aligned}
\ell^{U}_{l,i}
&=c_l(h_{l,i}), \\
\ell^{G}_{l,t,i}
&=c_l(h_{l,i})+e_{l,m(t)}(h_{l,i}).
\end{aligned}
\label{eq:core_expansion_formulation}
\end{equation}

Thus, understanding ranks tokens using only the shared component, whereas generation starts from the same core score and adds the active segment-specific residual.  The residual can raise or lower a token's rank as denoising progresses, and a larger final logit indicates higher priority for subsequent computation.

\subsection{Core-Expansion Router Optimization}
\label{sec:method_training}

Although Equation~\ref{eq:core_expansion_formulation} defines the additive core–expansion form, it does not ensure that the two components acquire their intended functional roles in a single training stage. 
We therefore use a generation-first, two-stage schedule motivated by the asymmetric transfer observed above. Generation first exposes the router to both recurring and progress-dependent importance, after which joint training anchors the core across tasks while preserving generation-specific residuals.  A straight-through estimator propagates gradients through the discrete top-$K$ routing decision.

\noindent\textbf{Stage I: Generation decomposition.}
We first train CE-Router using only generation samples, with timesteps drawn from all $M$ segments.  For a sample with $t\in\mathcal{T}_m$, the active path is $c_l+e_{l,m}$, and gradients update $\theta_l^{c}$ together with $\theta_{l,m}^{e}$.  Because the same core is activated for every segment, it receives supervision throughout the trajectory, whereas each expansion is updated only by its assigned segment.  This asymmetric exposure encourages the core to capture recurring patterns and the expansions to model progress-dependent deviations.

\noindent\textbf{Stage II: Cross-task core alignment.}
Starting from the Stage I parameters, we jointly train on understanding and generation samples.  Understanding activates only $c_l$ and directly supervises $\theta_l^{c}$.  A generation sample from segment $m$ retains the Stage I path, activating $c_l+e_{l,m}$ and updating both parameter sets. Joint training therefore aligns the core with the component useful to both
tasks while preserving segment-specific generation corrections in the expansions.

For a training sample from task $q\in\{U,G\}$, the router is optimized using the corresponding task objective and a dense-model consistency term:
\begin{equation}
\mathcal{L}
=
\mathcal{L}_{\mathrm{task}}^{q}
+
\lambda_{\mathrm{con}}
\left\|
z_{\mathrm{CE}}
-
z_{\mathrm{dense}}
\right\|_2^2,
\end{equation}
where $\mathcal{L}_{\mathrm{task}}^{q}$ is the original objective of task $q$, and $z_{\mathrm{CE}}$ and $z_{\mathrm{dense}}$ are the representations produced by routed and dense execution on the same sample, respectively.  In Stage I, $q=G$; in Stage II, $q$ follows the sampled task.  The consistency term limits representation drift introduced by routing, while the task objective teaches the router which computation can be removed without compromising task performance.

\subsection{Inference with Unified Computation Scheduling}
\label{sec:method_inference}

At inference, CE-Router serves two complementary roles.  It first converts the core--expansion logits into retained and dropped token sets for direct sequence compaction.  UCS then reuses the same decisions to coordinate acceleration across layers, channels, diffusion heads, and denoising steps.

\textbf{Task-adaptive token compaction.}
At routed layer $l$ and denoising timestep $t$, let $\ell_{l,t,i}$ denote the applicable understanding or generation logit from
Equation\,\ref{eq:core_expansion_formulation}.  We map it to a bounded importance score and retain the $K$ highest-scoring tokens:
\begin{equation}
\begin{aligned}
p_{l,t,i}
&=\operatorname{sigmoid}(\ell_{l,t,i}), \\
\mathcal{R}_{l,t}
&=\operatorname{TopK}
\left(\{p_{l,t,i}\}_{i=1}^{N},K\right).
\end{aligned}
\label{eq:router_inference_outputs}
\end{equation}

The remaining tokens form the dropped set $\mathcal{D}_{l,t}$.  Tokens in $\mathcal{R}_{l,t}$ are packed into a shorter sequence for attention and FFN computation, whereas $\mathcal{D}_{l,t}$ follows the bypass path. The timestep is omitted for understanding.

For multimodal understanding, routing is performed only during prefill and uses the core scorer $c_l$ without an expansion.  Compaction through the routed layers produces a compressed key--value cache.  CE-Router is then disabled during autoregressive decoding, which reuses this cache and avoids repeatedly attending to dropped visual tokens.

For visual generation, routing is recomputed at every denoising timestep. At $t\in\mathcal{T}_m$, CE-Router evaluates the active path $c_l+e_{l,m}$. The retained set can therefore evolve with denoising progress rather than remaining fixed throughout the trajectory.

\textbf{A unified signal for computation scheduling.}
The routing decisions provide two complementary signals.  Within a timestep, $\mathcal{R}_{l,t}$ identifies the tokens that spatial acceleration must preserve.  Across timesteps, overlap between drop sets indicates whether the redundancy pattern is stable enough for temporal reuse.  UCS uses the retained set to calibrate layer skipping and FFN pruning, and drop-set stability to control diffusion-head cache reuse and denoising-step early exit. We instantiate UCS through four coordinated operators, whose candidate actions are gated by operator-level thresholds.

\noindent\textbf{(1) Layer skipping.}
Training-free layer skipping commonly identifies redundant layers using the mean input--output cosine similarity over all tokens. This average can be dominated by already redundant tokens and conceal substantial updates to the tokens retained by CE-Router. For each candidate layer, we therefore compute the redundancy gap
\begin{equation}
    g_{l,t}
    =
    \bar{\rho}_{l,t}^{\mathrm{all}}
    -
    \bar{\rho}_{l,t}^{\mathrm{ret}},
\label{eq:redundancy_gap}
\end{equation}
where $\bar{\rho}_{l,t}^{\mathrm{all}}$ and $\bar{\rho}_{l,t}^{\mathrm{ret}}$ are the mean input--output cosine similarities over all tokens and over $\mathcal{R}_{l,t}$, respectively. We skip a candidate layer only if $g_{l,t}\leq\delta_{\mathrm{LS}}$.  A large positive gap indicates that the retained tokens change more than the sequence-level average and hence that the layer should still be executed.

\noindent\textbf{(2) FFN pruning.}
Following the activation-aware criterion of Wanda~\cite{sun2024wanda}, we estimate the contribution of each FFN channel from its activation on retained tokens and the magnitude of its output weights.  Let $z_{l,t,i,c}$ denote the intermediate activation of channel $c$ for retained token $i$.  We define its saliency as
\begin{equation}
    s_{l,t,c}
    =
    \frac{1}{|\mathcal{R}_{l,t}|}
    \sum_{i\in\mathcal{R}_{l,t}}
    |z_{l,t,i,c}|
    \left\|W^{\mathrm{down}}_{l,:,c}\right\|_2.
\label{eq:ffn_saliency}
\end{equation}

Let $\mathcal{P}_l$ be the set of channels proposed for removal by the static mask. Their saliency leakage is
\begin{equation}
    \lambda_{l,t}
    =
    \frac{
        \sum_{c\in\mathcal{P}_l}s_{l,t,c}
    }{
        \sum_{c}s_{l,t,c}
    }.
\label{eq:ffn_leakage}
\end{equation}

We execute the pruned FFN when $\lambda_{l,t}\leq\delta_{\mathrm{FFN}}$ and restore the full FFN otherwise.  Evaluating leakage on $\mathcal{R}_{l,t}$ preserves static pruning when the removed channels are unimportant to the retained tokens, while recovering full capacity when those channels carry task-relevant computation.

\noindent\textbf{(3) Diffusion-head cache reuse.}
Fixed-interval caching~\cite{ke2026flashu} reuses diffusion-head features without checking whether the underlying redundancy pattern has changed. For the generation-trajectory operators, we suppress the layer index and denote the router drop set at timestep $t$ by $\mathcal{D}_t$.  Let $t_{\mathrm{ref}}$ be the timestep at which the cached feature was last refreshed. We measure agreement between the current and reference drop sets using
\begin{equation}
    J_{t,t_{\mathrm{ref}}}
    =
    \frac{
        |\mathcal{D}_{t}\cap\mathcal{D}_{t_{\mathrm{ref}}}|
    }{
        |\mathcal{D}_{t}\cup\mathcal{D}_{t_{\mathrm{ref}}}|
    }.
\label{eq:drop_jaccard}
\end{equation}

A larger $J_{t,t_{\mathrm{ref}}}$ indicates that the router identifies a more stable redundancy pattern across the two timesteps.

Because raw overlap also increases with the drop ratio, we calibrate it by the overlap expected from random dropping with the same ratios:
\begin{equation}
    \mathrm{RDA}_{t,t_{\mathrm{ref}}}
    =
    \frac{
        J_{t,t_{\mathrm{ref}}}
        -
        J^{\mathrm{rand}}_{t,t_{\mathrm{ref}}}
    }{
        1-
        J^{\mathrm{rand}}_{t,t_{\mathrm{ref}}}
    },
\label{eq:rda}
\end{equation}
where $J^{\mathrm{rand}}_{t,t_{\mathrm{ref}}}$ is the corresponding expected Jaccard similarity.  We reuse the cached feature when $\mathrm{RDA}_{t,t_{\mathrm{ref}}}\geq\delta_{\mathrm{DC}}$ and refresh it otherwise. The cache horizon therefore adapts to the current sample and denoising stage, avoiding stale features when the routing pattern changes and unnecessary refreshes when it remains stable.

\noindent\textbf{(4) Denoising-step early exit.}
Adaptive sampling methods often require a separately trained confidence estimator~\cite{le2026dsa}.  UCS instead uses the temporal stability of CE-Router decisions as a convergence proxy.  Specifically, we measure the adjacent-step similarity $J_{t,t-1}$.  If it remains above $\delta_{\mathrm{EE}}$ for consecutive timesteps in the late denoising stage, we terminate the remaining iterations.

Rather than directly returning the current state, we approximate the skipped updates by projecting $x_{t+1}$ to the terminal solver time using the current velocity $v_t$:
\begin{equation}
    x_{\mathrm{final}}
    =
    x_{t+1}
    +
    v_t
    \left(
        \tau_{\mathrm{end}}-\tau_{t+1}
    \right),
\label{eq:terminal_projection}
\end{equation}
where $x_{t+1}$ is the current solver state at time $\tau_{t+1}$, $v_t$ is the current velocity prediction, and $\tau_{\mathrm{end}}$ is the terminal solver time.  This makes the denoising length sample-adaptive and removes unnecessary late-stage iterations without introducing an additional prediction head.

\begin{table*}[t]
    \centering
    
    \small
    \setlength{\tabcolsep}{3.8pt}
    \renewcommand{\arraystretch}{1.3}

    \resizebox{\linewidth}{!}{%
    \begin{tabular}{c|l|ccc|ccc|ccc}
        \toprule
        \multirow{2}{*}{\textbf{Scale}}
        & \multirow{2}{*}{\textbf{Method}}
        & \multicolumn{3}{c|}{\textbf{Understanding}}
        & \multicolumn{3}{c|}{\textbf{GenEval}}
        & \multicolumn{3}{c}{\textbf{DPG-Bench}} \\

        &
        & \textbf{Rel. Avg. (\%)} $\uparrow$
        & \textbf{FLOPs (T)} $\downarrow$
        & \textbf{Latency (ms)} $\downarrow$
        & \textbf{Score} $\uparrow$
        & \textbf{FLOPs (P)} $\downarrow$
        & \textbf{Latency (s)} $\downarrow$
        & \textbf{Score} $\uparrow$
        & \textbf{FLOPs (P)} $\downarrow$
        & \textbf{Latency (s)} $\downarrow$ \\
        \midrule

        \multirow{5}{*}{1.5B}
        & Show-o2
        & 100.00
        & 6.37
        & 474.41
        & 71.30
        & 0.99
        & 12.46
        & 85.25
        & 0.49
        & 5.35 \\

        & Show-o2 + FlashU
        & 83.19
        & 6.27
        & 483.85
        & 64.16
        & \underline{0.78}
        & \underline{11.37}
        & 81.06
        & \underline{0.37}
        & \underline{5.00} \\

        & Show-o2 + UniMoD
        & 95.04
        & 5.42
        & 461.52
        & 62.83
        & 0.84
        & 12.22
        & 81.72
        & 0.42
        & 5.14 \\

        \rowcolor{gray!10}
        \cellcolor{white}
        & Show-o2 + CE-Router (Ours)
        & \textbf{99.36}
        & \underline{3.04}
        & \underline{300.32}
        & \textbf{66.33}
        & 0.85
        & 12.11
        & \textbf{82.86}
        & 0.42
        & 5.15 \\

        \rowcolor{gray!10}
        \cellcolor{white}
        & Show-o2 + CE-Router + UCS
        & \underline{98.03}
        & \textbf{2.73}
        & \textbf{245.25}
        & \underline{64.45}
        & \textbf{0.70}
        & \textbf{10.71}
        & \underline{81.85}
        & \textbf{0.33}
        & \textbf{4.75} \\

        \midrule
        
        \multirow{5}{*}{7B}
        & BAGEL
        & 100.00
        & 25.32
        & 487.68
        & 78.80
        & 10.60
        & 81.17
        & 83.95
        & 5.25
        & 40.01 \\

        & BAGEL + FlashU
        & 94.16
        & \underline{22.47}
        & 465.48
        & 72.48
        & \underline{7.35}
        & \underline{77.21}
        & 79.59
        & \underline{3.64}
        & \underline{37.18} \\

        & BAGEL + UniMoD
        & 96.29
        & 23.58
        & 458.69
        & 54.16
        & 9.83
        & 80.98
        & 80.72
        & 4.86
        & 39.34 \\

        \rowcolor{gray!10}
        \cellcolor{white}
        & BAGEL + CE-Router (Ours)
        & \textbf{99.96}
        & 23.31
        & \underline{417.76}
        & \textbf{75.92}
        & 10.07
        & 81.01
        & \textbf{83.49}
        & 4.89
        & 39.30 \\

        \rowcolor{gray!10}
        \cellcolor{white}
        & BAGEL + CE-Router + UCS
        & \underline{96.44}
        & \textbf{17.27}
        & \textbf{403.34}
        & \underline{73.15}
        & \textbf{7.29}
        & \textbf{75.99}
        & \underline{80.77}
        & \textbf{3.60}
        & \textbf{36.11} \\

        \bottomrule
    \end{tabular}%
    }
    \caption{
        Comparison of performance and inference efficiency on multimodal
        understanding and image generation.
        UCS denotes our Unified Computation Scheduling strategy.
        \textbf{Bold} indicates the best result.
        {\underline{Underlining}} indicates the second-best result.
    }
    \label{tab:main_results}
\end{table*}

\section{Experiments}

\subsection{Experimental Setup}
\textbf{Models and Baselines.}
We evaluate CE-Router on two architecturally distinct UMM backbones: the shared-Transformer Show-o2-1.5B~\cite{xie2026showo2} and the Mixture-of-Transformers BAGEL-7B-MoT~\cite{deng2025bagel}.
On each backbone, we compare CE-Router with FlashU~\cite{ke2026flashu}, UniMoD~\cite{mao2025unimod} and the dense backbone.
All methods are evaluated under matched inference protocols.

\textbf{CE-Router Training.}
We set $M=4$ and use two-stage training on both backbones.
On Show-o2, the training data are aligned with the Show-o2 setup~\cite{xie2026showo2}: Stage I uses 10K generation samples, while Stage II adds 10K understanding samples at a $3{:}1$ generation-to-understanding ratio; each stage is trained for 1920 steps.
On BAGEL, we scale the generation set to 100K, retain 10K understanding samples in Stage II at the same ratio, and train each stage for 4K steps.
Show-o2 trains CE-Router with LoRA on the frozen backbone, whereas BAGEL directly trains the router.
Full details are provided in Appendix A.2.

\textbf{UCS Settings.}
UCS builds on FlashU's task-specific operators with slightly enlarged candidate budgets, while gating each candidate action with the CE-Router signals. Additional thresholds are calibrated on a small held-out set and reported in the Appendix A.4.

\textbf{Evaluation.}
We evaluate multimodal understanding on 7 benchmarks~\cite{kembhavi2016ai2d,hudson2019gqa,liu2024mmbench,fu2025mme,yue2024mmmu,chen2024mmstar,li2024seedbench}, and image generation on GenEval and DPG-Bench~\cite{ghosh2023geneval,hu2024ella}, at a resolution of $432\times432$, using 100 and 50 sampling steps respectively. Additional experimental details are provided in the Appendix A.1.

\subsection{Experimental Results}
\textbf{Quantitative Comparison.}
\label{sec:task_results}
Table~\ref{tab:main_results} shows that CE-Router achieves the best overall results on both backbones. It preserves 99.36\% and 99.96\% of the original understanding scores on Show-o2 and BAGEL respectively, while attaining the highest GenEval and DPG-Bench scores. These results highlight the consistent effectiveness of CE-Router across architectures and tasks.
With UCS, CE-Router further reduces FLOPs and latency across all tasks and still leads other accelerated methods.

\begin{table*}[t]
    \centering

    \small
    \setlength{\tabcolsep}{3.8pt}
    \renewcommand{\arraystretch}{0.8}

    \begin{tabular*}{\textwidth}{
        @{\extracolsep{\fill}}
        l|ccc|c
        @{}
    }
        \toprule
        \textbf{Setting}
        & \textbf{UMM Avg. (\%)} $\uparrow$
        & \textbf{GenEval} $\uparrow$
        & \textbf{DPG-Bench} $\uparrow$
        & \textbf{Router Params. (M)} $\downarrow$ \\
        \midrule

        \rowcolor{gray!12}
        \multicolumn{5}{l}{\textit{Router Architecture}} \\

        Fully Shared
        & 89.94
        & 59.60
        & 77.62
        & \textbf{113.72} \\

        Independent
        & \underline{99.42}
        & \textbf{66.61}
        & \textbf{83.15}
        & 231.43 \\

        Shared Core + Und. Exp
        & \textbf{99.52}
        & 61.28
        & 76.13
        & \underline{185.15} \\

        Shared Core + Gen. Exp (Ours)
        & 99.36
        & \underline{66.33}
        & \underline{82.86}
        & \underline{185.15} \\

        \midrule

        \rowcolor{gray!12}
        \multicolumn{5}{l}{\textit{Number of Generation Segments}} \\

        $M=1$
        & \underline{99.32}
        & 60.29
        & 76.19
        & \textbf{115.76} \\

        $M=2$
        & 99.15
        & 63.78
        & 80.11
        & \underline{138.86} \\

        $M=4$ (Ours)
        & \textbf{99.36}
        & \textbf{66.33}
        & \textbf{82.86}
        & 185.15 \\

        $M=8$
        & 99.29
        & \underline{66.31}
        & \underline{82.47}
        & 277.72 \\

        \midrule

        \rowcolor{gray!12}
        \multicolumn{5}{l}{\textit{Training Strategy}} \\

        Joint UMM + T2I
        & \underline{98.09}
        & 65.83
        & \underline{83.07}
        & 185.15 \\

        UMM $\rightarrow$ UMM + T2I
        & 95.61
        & \textbf{67.06}
        & \textbf{83.37}
        & 185.15 \\

        T2I $\rightarrow$ UMM + T2I (Ours)
        & \textbf{99.36}
        & \underline{66.33}
        & 82.86
        & 185.15 \\

        \midrule

        \rowcolor{gray!12}
        \multicolumn{5}{l}{
            \textit{Consistency-Loss Weight $\lambda_{\mathrm{con}}$}
        } \\

        $0$ (w/o Consistency Loss)
        & 93.03
        & 52.74
        & 65.95
        & 185.15 \\

        $0.1$
        & 98.12
        & 64.72
        & 82.34
        & 185.15 \\

        $0.25$ (Ours)
        & \textbf{99.36}
        & \textbf{66.33}
        & \textbf{82.86}
        & 185.15 \\

        $0.5$
        & \underline{98.78}
        & \underline{66.25}
        & \underline{82.79}
        & 185.15 \\

        \bottomrule
    \end{tabular*}

    \caption{
        Ablation of the router architecture, generation segmentation,
        training strategy, and consistency-loss weight.
        \textbf{Bold} and \underline{underlined} denotes the best and
        second-best results.
    }
    \label{tab:router_ablation}
\end{table*}

\newcommand{\dropcell}[2]{%
    #1%
    \raisebox{-0.55ex}{%
        \hspace{0.16em}%
        {\scriptsize$\downarrow\,#2$}%
    }%
}

\begin{table*}[!t]
    \centering

    \footnotesize
    \setlength{\tabcolsep}{4.5pt}
    \renewcommand{\arraystretch}{1.05}

\begin{tabularx}{\textwidth}{
    @{}
    >{\raggedright\arraybackslash}p{0.22\textwidth}
    !{\hspace{0.45em}\vrule width 0.45pt\hspace{0.45em}}
    >{\centering\arraybackslash}X
    >{\centering\arraybackslash}X
    >{\centering\arraybackslash}X
    @{}
}
        \toprule

        \multicolumn{4}{l}{
            \textit{(a) Multimodal Understanding}
        } \\
        \cmidrule(lr){1-4}

        \textbf{Configuration}
        & \textbf{UMM Avg. (\%)} $\uparrow$
        & \textbf{TFLOPs} $\downarrow$
        & \textbf{Latency (ms)} $\downarrow$ \\
        \midrule

        \rowcolor{gray!10}
        CE-Router
        & \textbf{99.36}
        & 3.04
        & 300.32 \\

        \hspace{0.6em}$+$ FlashU
        & \dropcell{80.62}{18.74}
        & \dropcell{\underline{2.82}}{0.22}
        & \dropcell{259.12}{41.20} \\

        \rowcolor{gray!10}
        \hspace{0.6em}$+$ UCS
        & \dropcell{98.03}{1.33}
        & \dropcell{\textbf{2.73}}{0.31}
        & \dropcell{\textbf{245.25}}{55.07} \\

        \hspace{1.2em}\textit{w/o} Layer Skipping
        & \dropcell{\underline{98.72}}{0.64}
        & \dropcell{2.97}{0.07}
        & \dropcell{282.00}{18.32} \\

        \hspace{1.2em}\textit{w/o} FFN Pruning
        & \dropcell{98.48}{0.88}
        & \dropcell{2.83}{0.21}
        & \dropcell{\underline{256.02}}{44.30} \\

        \midrule

        \multicolumn{4}{l}{
            \textit{(b) Text-to-Image Generation}
        } \\
        \cmidrule(lr){1-4}

        \textbf{Configuration}
        & \textbf{DPG-Bench} $\uparrow$
        & \textbf{TFLOPs} $\downarrow$
        & \textbf{Latency (s)} $\downarrow$ \\
        \midrule

        \rowcolor{gray!10}
        CE-Router
        & \textbf{82.86}
        & 415.64
        & 5.15 \\

        \hspace{0.6em}$+$ FlashU
        & \dropcell{79.58}{3.28}
        & \dropcell{370.92}{44.72}
        & \dropcell{4.97}{0.18} \\

        \rowcolor{gray!10}
        \hspace{0.6em}$+$ UCS
        & \dropcell{81.85}{1.01}
        & \dropcell{\textbf{334.44}}{81.20}
        & \dropcell{\textbf{4.75}}{0.40} \\

        \hspace{1.2em}\textit{w/o} Layer Skipping
        & \dropcell{82.04}{0.82}
        & \dropcell{373.76}{41.88}
        & \dropcell{4.82}{0.33} \\

        \hspace{1.2em}\textit{w/o} FFN Pruning
        & \dropcell{81.88}{0.98}
        & \dropcell{\underline{339.16}}{76.48}
        & \dropcell{\underline{4.76}}{0.39} \\

        \hspace{1.2em}\textit{w/o} Diffusion Cache
        & \dropcell{\underline{82.69}}{0.17}
        & \dropcell{378.92}{36.72}
        & \dropcell{4.92}{0.23} \\

        \hspace{1.2em}\textit{w/o} Early Exit
        & \dropcell{82.42}{0.44}
        & \dropcell{380.87}{34.77}
        & \dropcell{4.98}{0.17} \\

        \bottomrule
    \end{tabularx}

    \caption{
        Comparison and component ablation of Unified Computation Scheduling
        (UCS) on multimodal understanding and text-to-image generation.
    }
    \label{tab:ucs_ablation}
\end{table*}

\begin{figure}[t]
  \centering
  \includegraphics[width=\columnwidth]{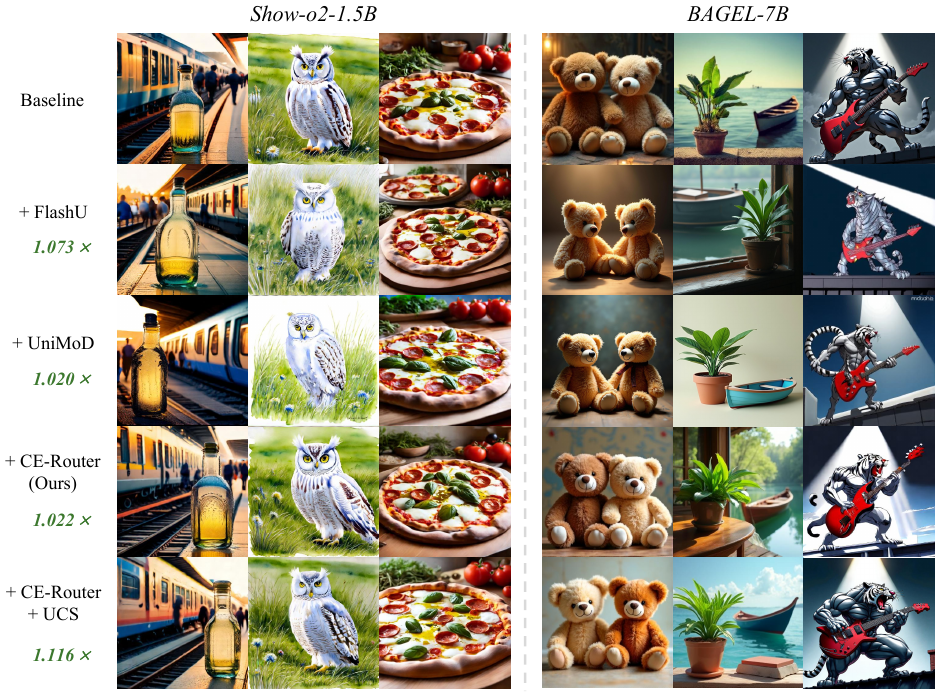}
    \caption{Qualitative results on GenEval and DPG-Bench.
    }
\label{fig:qualitative_results}
\end{figure}

\textbf{Qualitative Comparison.}
Figure~\ref{fig:qualitative_results} presents qualitative results across diverse scenes, including outdoor, hand-painted, still-life, and cartoon examples. 
FlashU and UniMoD produce object deformations and inconsistent colors, including a distorted bottle and a washed-out owl, whereas CE-Router preserves cleaner object geometry, more consistent colors, and richer details.
CE-Router also better preserves realism and spatial relations, as reflected in the tiger's natural pose and the coherent arrangement of the plant and boat. These advantages are largely retained after applying UCS. More results are provided in the Appendix D.

\textbf{Efficiency and Latency.}
CE-Router best preserves performance, while UCS provides the corresponding runtime gains. CE-Router alone can be slower than FlashU, especially for generation, since FlashU compresses multiple components beyond token computation. 
With UCS, the full method closes this latency gap while retaining higher accuracy. On Show-o2 GenEval, it achieves a score of 64.45 in 10.71\,s, compared with 64.16 in 11.37\,s for FlashU.
Overall, CE-Router preserves quality, while UCS converts its dynamic sparsity into practical acceleration, yielding the strongest performance--latency trade-off.

\subsection{Ablation Studies}
\label{sec:ablation}

\textbf{Router Parameterization.}
Table~\ref{tab:router_ablation} compares four router parameterizations on Show-o2. 
A fully shared router uses the fewest parameters but markedly degrades both understanding and generation, whereas independent routers achieve the highest generation scores at the cost of substantial parameter overhead.
Under the same parameter budget, the understanding-expanded variant improves Rel. Avg. by only 0.16 points over our design but trails it by 5.05 points on GenEval and 6.73 points on DPG-Bench.
This indicates that task-specific expansion brings only marginal gains to understanding but is substantially more beneficial to generation, supporting our core-shared, generation-expanded design.

\textbf{Generation Segmentation.}
As shown in Table~\ref{tab:router_ablation}, increasing $M$ from 1 to 4 steadily improves GenEval and DPG-Bench while keeping understanding performance stable, but increases the router parameter count. 
At $M=8$, generation performance saturates even as the parameter count rises to 277.72M. We therefore use $M=4$, which delivers the best overall results with 185.15M parameters while remaining more compact than independent routing.

\textbf{Training Configuration.}
Table~\ref{tab:router_ablation} shows that joint UMM--T2I training is stable but suboptimal, while UMM-first training improves T2I at the cost of reducing UMM to 95.61\%. Our generation-first strategy achieves the best overall balance. For the consistency objective, $\lambda_{\mathrm{con}}=0.25$ achieves the best overall performance. Smaller or larger weights are less effective, while removing the consistency loss causes substantial degradation on both tasks.

\textbf{Unified Computation Scheduling.}
Table~\ref{tab:ucs_ablation} evaluates how effectively UCS extends CE-Router from token routing to operator-level acceleration on Show-o2. Compared with applying FlashU directly to CE-Router, UCS achieves larger reductions in computation and latency, while limiting the understanding and generation score drops to 1.33 and 1.01 points, far below FlashU's 18.74- and 3.28- point drops. This advantage stems from its trained, unified routing signal, which captures task-dependent redundancy along the computation trajectory and avoids overly aggressive operations. By contrast, FlashU relies on independently designed, module-specific heuristics that are not calibrated to the model's actual redundancy, resulting in a less favorable accuracy--efficiency trade-off.

\section{Limitations and Future Work}
Our current study focuses on transformer-based unified multimodal models that process visual representations as token sequences. Extending CE-Router to models with different modality interfaces or generation paradigms may require architecture-specific adaptations. Although our current implementation already achieves end-to-end latency reductions, the magnitude of practical speedup can vary across hardware and software stacks due to differences in kernel and runtime efficiency for dynamic token compaction and operator-level scheduling. Future work will investigate hardware-aware scheduling and optimized kernels for these dynamic operations to improve portability and further enhance deployment efficiency.

\section{Conclusion}
In this work, we investigate how computational redundancy is shared between understanding and generation in unified multimodal models. Our analysis reveals an asymmetric core-expansion structure: a stable importance component is transferable across tasks, while generation requires progress-conditioned corrections. CE-Router captures this structure with a task-shared core scorer and segment-conditioned generation expansions, enabling adaptive token compaction. Building on the same learned routing signal, Unified Computation Scheduling coordinates layer skipping, FFN pruning, diffusion-head cache reuse, and denoising-step early exit. Experiments on Show-o2 and BAGEL show consistent quality--efficiency gains across both tasks, including a
1.93$\times$ end-to-end speedup while retaining 98.03\% of dense understanding performance. These results establish cross-task redundancy modeling as an effective foundation for efficient unified multimodal inference.


\bibliographystyle{plainnat}
\bibliography{main}

\clearpage
\newpage

\beginappendix

\setcounter{secnumdepth}{2}

\section{Experimental Details}
\label{sec:app_experimental_details}

The main paper describes the evaluated models and tasks, the core design of CE-Router, and the four UCS operators. Rather than repeating the methodological derivations, this section provides the training configurations, baseline implementations, UCS operating points, and efficiency measurement protocol.

\subsection{Evaluation and Efficiency Measurement Protocol}
\label{sec:app_eval_protocol}

The main paper evaluates two backbones, Show-o2-1.5B~\cite{xie2026showo2} and BAGEL-7B-MoT~\cite{deng2025bagel}, on understanding and generation tasks. Table~\ref{tab:app_benchmarks} summarizes the benchmarks, generation resolution, and number of denoising steps~\cite{kembhavi2016ai2d,fu2025mme,liu2024mmbench,hudson2019gqa,chen2024mmstar,li2024seedbench,yue2024mmmu,ghosh2023geneval,hu2024ella}.

All generation training, performance evaluation, and efficiency measurements use a resolution of $432\times432$. TFLOPs and latency are measured over the complete end-to-end inference path rather than at a single layer or denoising step. Table~\ref{tab:app_efficiency_protocol} details the sample-selection and averaging procedures.

\begin{table*}[!b]
    \centering
    \footnotesize
    \setlength{\tabcolsep}{5pt}
    \renewcommand{\arraystretch}{1.15}

 \begin{tabularx}{\textwidth}{
        @{}
        >{\raggedright\arraybackslash}m{0.10\textwidth}
        >{\raggedright\arraybackslash}X
        >{\centering\arraybackslash}m{0.28\textwidth}
        >{\centering\arraybackslash}m{0.30\textwidth}
        @{}
    }   
        \toprule

        \textbf{Task}
        & \textbf{Benchmark}
        & \textbf{Resolution}
        & \textbf{Denoising Steps} \\
        \midrule

        Und.
        &
        \begin{tabular}[c]{@{}l@{}}
            AI2D, MME, MMBench, GQA,\\
            MMStar, SEEDBench, MMMU
        \end{tabular}
        & -- 
        & -- \\

        \midrule

        \multirow{2}{*}{Gen.}
        & GenEval
        & $432 \times 432$
        & 100 \\

        & DPG-Bench
        & $432 \times 432$
        & 50 \\

        \bottomrule
    \end{tabularx}

    \caption{Evaluation benchmarks and generation inference protocol.}
    \label{tab:app_benchmarks}
\end{table*}

\begin{table*}[!b]
    \centering

    \footnotesize
    \setlength{\tabcolsep}{5pt}
    \renewcommand{\arraystretch}{1.15}

    \begin{tabularx}{\textwidth}{
        @{}
        >{\raggedright\arraybackslash}p{0.075\textwidth}
        >{\raggedright\arraybackslash}p{0.135\textwidth}
        >{\raggedright\arraybackslash}p{0.390\textwidth}
        >{\raggedright\arraybackslash}X
        @{}
    }
        \toprule

        \textbf{Task}
        & \textbf{Evaluation Set}
        & \textbf{Sample Protocol}
        & \textbf{Measurement Boundary} \\
        \midrule

        Und.
        & MME
        & Discard the first and last $1/16$ of samples in inference order
          and average over the middle $7/8$
        & Complete end-to-end inference for one sample \\

        \midrule

        \multirow[t]{2}{*}{Gen.}
        & GenEval
        & Use 100 samples, discard the first and last 2, and average over
          the middle 96
        & Complete 100-step denoising trajectory for one sample \\

        \addlinespace[2pt]

        & DPG-Bench
        & Use 100 samples, discard the first and last 2, and average over
          the middle 96
        & Complete 50-step denoising trajectory for one sample \\

        \bottomrule
    \end{tabularx}

    \caption{Protocol for measuring end-to-end TFLOPs and latency.}
    \label{tab:app_efficiency_protocol}
\end{table*}

\subsection{CE-Router Training Details}
\label{sec:app_cerouter_training}

\subsubsection{General Configuration and Keep-Ratio Schedule}

The main paper introduces Core-Expansion Scoring and the two-stage optimization procedure. For reproducibility, Table~\ref{tab:app_cerouter_general} provides the common router configuration and keep-ratio schedule used in the two stages.

\begin{table*}[!ht]
    \centering
    \footnotesize
    \setlength{\tabcolsep}{3pt}
    \renewcommand{\arraystretch}{1.10}

    \begin{tabularx}{\textwidth}{
        @{}
        >{\raggedright\arraybackslash}p{0.48\textwidth}
        >{\raggedright\arraybackslash}X
        @{}
    }
        \toprule
        \textbf{Item} & \textbf{Setting} \\
        \midrule

        Gen. segments / expansion scorers $M$
        & 4 \\

        Discrete routing gradient
        & Straight-through estimator \\

        \addlinespace[2pt]

        Stage I keep ratio
        & Fixed at $r = K/N = 0.8$ \\

        Stage II keep ratio
        & Sampled from the predefined training interval $[0.6, 0.95]$ \\

        Reported inference keep ratio
        & $r = 0.8$ \\

        \addlinespace[2pt]

        Gen. resolution
        & $432 \times 432$ \\

        \bottomrule
    \end{tabularx}

    \caption{Common training configuration of CE-Router.}
    \label{tab:app_cerouter_general}
\end{table*}

Stage I fixes $r=0.8$ to reduce the risk of non-convergence caused by discrete routing early in training. Stage II instead samples the keep ratio from the predefined interval, encouraging the router to learn a stable token ranking rather than adapt only to one fixed top-$K$ set. The same router can therefore be directly used at different keep ratios within the training interval. All reported results use $r=0.8$, placing CE-Router and methods such as FlashU~\cite{ke2026flashu} at similar TFLOPs operating points and enabling task performance to be compared under approximately matched computation. This strategy is used for both Show-o2 and BAGEL.

\begin{table*}[!ht]
    \centering
    \footnotesize
    \setlength{\tabcolsep}{6pt}
    \renewcommand{\arraystretch}{1.12}

    \begin{tabularx}{\textwidth}{
        @{}
        >{\raggedright\arraybackslash}p{0.21\textwidth}
        >{\raggedright\arraybackslash}X
        >{\raggedright\arraybackslash}X
        @{}
    }
        \toprule
        \textbf{Setting}
        & \textbf{Stage I: Generation Decomposition}
        & \textbf{Stage II: Cross-Task Core Alignment} \\
        \midrule

        Gen. data
        & 10K, aligned with Show-o2~\cite{xie2026showo2}
        & Same 10K training data \\
    
        Und. data
        & --
        & 10K \\
    
        Und. composition
        & --
        & Aligned with Show-o2~\cite{xie2026showo2} \\

        Task sampling
        & Generation only
        & Gen./Und. $= 3:1$ \\

        \addlinespace[2pt]

        Training steps
        & 1,920
        & 1,920 \\

        Router learning rate
        & $5 \times 10^{-5}$
        & $5 \times 10^{-5}$ \\

        LoRA learning rate
        & $1 \times 10^{-5}$
        & $1 \times 10^{-5}$ \\

        Optimizer
        & AdamW
        & AdamW \\

        Consistency weight $\lambda_{\mathrm{con}}$
        & 0.25
        & 0.25 \\

        \addlinespace[2pt]

        Keep-ratio schedule
        & Fixed at 0.8
        & Sampled from $[0.6, 0.95]$ \\

        Initialization
        & Show-o2 checkpoint
        & Stage I checkpoint \\

        Resolution
        & $432 \times 432$
        & $432 \times 432$ \\

        \bottomrule
    \end{tabularx}

    \caption{Two-stage CE-Router training configuration on Show-o2.}
    \label{tab:app_showo_cerouter_training}
\end{table*}

\subsubsection{CE-Router on Show-o2}

Table~\ref{tab:app_showo_cerouter_training} gives the complete configuration of the two training stages on Show-o2. The main paper explains the functional roles of the two stages; here, we provide only the data, optimization, and initialization settings.

The 10K understanding samples in Stage II are aligned with the training-data setup of Show-o2~\cite{xie2026showo2}. During training, generation and understanding samples are drawn at a $3{:}1$ ratio. At each training iteration, one of the four generation segments is randomly selected, with sampling probabilities of $1/3$, $1/3$, $1/6$, and $1/6$, respectively.

Both Show-o2 stages train the task-shared core scorer, four generation expansion scorers, and LoRA~\cite{hu2022lora} parameters. When computing the dense-model consistency target, CE-Router and LoRA are disabled, and the dense forward representation of the same sample is used. Table~\ref{tab:app_showo_lora} lists the detailed LoRA configuration.

The router and LoRA use separate optimizer parameter groups, with learning rates of $5\times10^{-5}$ and $1\times10^{-5}$, respectively.

\begin{table*}[!ht]
    \centering
    \footnotesize
    \setlength{\tabcolsep}{3pt}
    \renewcommand{\arraystretch}{1.10}

    \begin{tabularx}{\textwidth}{
        @{}
        >{\raggedright\arraybackslash}p{0.36\textwidth}
        >{\raggedright\arraybackslash}X
        @{}
    }
        \toprule
        \textbf{Parameter} & \textbf{Setting} \\
        \midrule

        Rank
        & 16 \\

        Alpha
        & 32 \\

        Scaling $\alpha/r$
        & 2 \\

        Dropout
        & 0.05 \\

        Task type
        & CAUSAL\_LM \\

        LoRA learning rate
        & $1 \times 10^{-5}$ \\

        \bottomrule
    \end{tabularx}

    \caption{LoRA configuration on Show-o2.}
    \label{tab:app_showo_lora}
\end{table*}

\begin{table*}[!t]
    \centering
    \scriptsize
    \setlength{\tabcolsep}{4pt}
    \renewcommand{\arraystretch}{1.10}

    \begin{tabularx}{\textwidth}{
        @{}
        >{\raggedright\arraybackslash}p{0.27\textwidth}
        >{\raggedright\arraybackslash}p{0.31\textwidth}
        >{\raggedright\arraybackslash}X
        @{}
    }
        \toprule

        \textbf{Setting}
        & \textbf{Stage I:} \textbf{Gen. decomposition}
        & \textbf{Stage II:} \textbf{Cross-task core alignment} \\
        \midrule

        Gen. data
        & 100K
        & 100K \\

        Und. data
        & --
        & 10K \\

        Task sampling
        & Gen. only
        & Gen./Und. $= 3:1$ \\

        \addlinespace[2pt]

        Training steps
        & 4K
        & 4K \\

        Router learning rate
        & $3 \times 10^{-5}$
        & $1.5 \times 10^{-5}$ \\

        Consistency weight
        & $\lambda_{\mathrm{con}} = 0.5$
        & $\lambda_{\mathrm{con}} = 0.75$ \\

        \addlinespace[2pt]

        Keep-ratio schedule
        & Fixed at 0.8
        & Sampled from $[0.6, 0.95]$ \\

        Trainable scorers
        & Core + expansion
        & Core + expansion \\

        \addlinespace[2pt]

        Initialization
        & BAGEL checkpoint
        & Stage I checkpoint \\

        Resolution
        & $432 \times 432$
        & $432 \times 432$ \\

        \bottomrule
    \end{tabularx}

    \caption{Two-stage CE-Router training configuration on BAGEL.}
    \label{tab:app_bagel_cerouter_training}
\end{table*}

\subsubsection{CE-Router on BAGEL}

BAGEL follows the same two-stage procedure, but expands the generation data to 100K and does not use LoRA. Table~\ref{tab:app_bagel_cerouter_training} summarizes its training configuration.

\subsection{FlashU Baseline Configuration}
\label{sec:app_flashu}

FlashU is applied directly to the dense backbone and uses neither additional training data nor trainable parameters. The shared resolution, sampling steps, and efficiency measurement protocol follow Section~\ref{sec:app_eval_protocol}; this section reports only the acceleration operators and operating points that affect the comparison.

\subsubsection{FlashU on Show-o2}

Table~\ref{tab:app_flashu_showo} summarizes the FlashU
configuration used on Show-o2. FFN pruning and layer skipping
are applied to both understanding and generation, whereas
diffusion-head cache reuse is applied only to generation.
For matched-step evaluation, fixed denoising-step reduction is
disabled for FlashU, which therefore executes the full 100-step
GenEval and 50-step DPG-Bench trajectories.

\begin{table*}[!ht]
    \centering
    \footnotesize
    \setlength{\tabcolsep}{3pt}
    \renewcommand{\arraystretch}{1.10}

    \begin{tabularx}{\textwidth}{
        @{}
        >{\raggedright\arraybackslash}p{0.34\textwidth}
        >{\raggedright\arraybackslash}X
        >{\centering\arraybackslash}p{0.25\textwidth}
        @{}
    }
        \toprule
        \textbf{Operator}
        & \textbf{Parameter}
        & \textbf{Setting} \\
        \midrule

        \rowcolor{gray!12}
        \multicolumn{3}{@{}l}{\textit{Understanding and Generation}} \\[-1pt]

        FFN pruning
        & FFN prune ratio
        & 20\% \\

        Layer skipping
        & Candidate layer-skip ratio
        & 20\% \\
        \hline

        \addlinespace[3pt]

        \rowcolor{gray!12}
        \multicolumn{3}{@{}l}{\textit{Generation only}} \\[-1pt]

        Diffusion-head cache
        & Fixed cache horizon
        & 5 steps \\

        \bottomrule
    \end{tabularx}

    \caption{FlashU operating points on Show-o2.}
    \label{tab:app_flashu_showo}
\end{table*}

\subsubsection{FlashU on BAGEL}

Table~\ref{tab:app_flashu_bagel} separately reports the understanding and generation configurations on BAGEL. Diffusion-head caching is disabled for BAGEL generation.

\begin{table*}[!ht]
    \centering
    \footnotesize
    \setlength{\tabcolsep}{3pt}
    \renewcommand{\arraystretch}{1.10}

    \begin{tabularx}{\textwidth}{
        @{}
        >{\raggedright\arraybackslash}p{0.34\textwidth}
        >{\raggedright\arraybackslash}X
        >{\centering\arraybackslash}p{0.25\textwidth}
        @{}
    }
        \toprule
        \textbf{Operator}
        & \textbf{Parameter}
        & \textbf{Setting} \\
        \midrule

        \rowcolor{gray!12}
        \multicolumn{3}{@{}l}{\textit{Understanding}} \\[-1pt]

        Visual-token pruning
        & Visual-token prune ratio
        & 15\% \\

        FFN pruning
        & FFN prune ratio
        & 12\% \\

        Layer skipping
        & Candidate layer-skip ratio
        & 10\% \\
        \hline

        \addlinespace[3pt]

        \rowcolor{gray!12}
        \multicolumn{3}{@{}l}{\textit{Generation}} \\[-1pt]

        FFN pruning
        & FFN prune ratio
        & 10\% \\

        Layer skipping
        & Candidate layer-skip ratio
        & 8\% \\

        Diffusion-head cache
        & Status
        & Disabled \\

        \bottomrule
    \end{tabularx}

    \caption{FlashU operating points on BAGEL.}
    \label{tab:app_flashu_bagel}
\end{table*}

\subsection{Unified Computation Scheduling}
\label{sec:app_ucs}

The main paper defines the four UCS operators and their equations. This section specifies only the correspondence between FlashU and UCS, the actual operating points, and the threshold-selection procedure. Following the notation in the main paper, the thresholds for layer skipping, FFN pruning, diffusion-head caching, and denoising-step early exit are denoted by $\delta_{\mathrm{LS}}$, $\delta_{\mathrm{FFN}}$, $\delta_{\mathrm{DC}}$, and $\delta_{\mathrm{EE}}$, respectively.

\subsubsection{FlashU-to-UCS Mapping}

UCS uses the task-specific operators corresponding to those in FlashU but permits slightly larger candidate budgets. The retained- and dropped-token signals from CE-Router then block potentially destructive operations. Table~\ref{tab:app_flashu_ucs_mapping} summarizes the mapping between the four corresponding operations.

\begin{table*}[!ht]
    \centering
    \footnotesize
    \setlength{\tabcolsep}{3pt}
    \renewcommand{\arraystretch}{1.10}

    \begin{tabularx}{\textwidth}{
        @{}
        >{\raggedright\arraybackslash}p{0.27\textwidth}
        >{\raggedright\arraybackslash}X
        @{}
    }
        \toprule
        \textbf{Component}
        & \textbf{Description} \\
        \midrule

        \rowcolor{gray!12}
        \multicolumn{2}{@{}l}{\textit{Layer Skipping}} \\[-1pt]

        FlashU candidate
        & Select candidate layers according to full-sequence redundancy \\

        UCS decision
        & Skip only when the retained-token redundancy gap satisfies
          $g_{l,t} \leq \delta_{\mathrm{LS}}$ \\
          \hline

        \addlinespace[3pt]

        \rowcolor{gray!12}
        \multicolumn{2}{@{}l}{\textit{FFN Pruning}} \\[-1pt]

        FlashU candidate
        & Use a static channel-pruning mask \\

        UCS decision
        & Execute the pruned FFN only when retained-token saliency leakage
          satisfies $\lambda_{l,t} \leq \delta_{\mathrm{FFN}}$ \\
          \hline

        \addlinespace[3pt]

        \rowcolor{gray!12}
        \multicolumn{2}{@{}l}{\textit{Diffusion-Head Cache}} \\[-1pt]

        FlashU candidate
        & Reuse features with a fixed horizon \\

        UCS decision
        & Adaptively reuse or refresh according to
          $\mathrm{RDA}_{t,t_{\mathrm{ref}}}$ and $\delta_{\mathrm{DC}}$ \\
          \hline

        \addlinespace[3pt]

        \rowcolor{gray!12}
        \multicolumn{2}{@{}l}{\textit{Denoising-Step Early Exit}} \\[-1pt]

        FlashU candidate
        & Reduce denoising steps by a fixed amount \\

        UCS decision
        & Terminate early according to late-stage router-mask similarity
          and $\delta_{\mathrm{EE}}$ \\

        \bottomrule
    \end{tabularx}

    \caption{Correspondence between FlashU operators and UCS decision signals.}
    \label{tab:app_flashu_ucs_mapping}
\end{table*}

\subsubsection{UCS on Show-o2}

Table~\ref{tab:app_ucs_showo} presents the FlashU operating points and UCS configuration side by side on Show-o2. UCS enlarges the layer-skipping candidate budget and replaces fixed caching and fixed step reduction with adaptive decisions controlled by routing stability.

\begin{table*}[!ht]
    \centering
    \scriptsize
    \setlength{\tabcolsep}{3pt}
    \renewcommand{\arraystretch}{1.10}

    \begin{tabularx}{\textwidth}{
        @{}
        >{\raggedright\arraybackslash}p{0.28\textwidth}
        >{\raggedright\arraybackslash}p{0.24\textwidth}
        >{\raggedright\arraybackslash}X
        @{}
    }
        \toprule
        \textbf{Operator}
        & \textbf{FlashU}
        & \textbf{CE-Router + UCS} \\
        \midrule

        \rowcolor{gray!12}
        \multicolumn{3}{@{}l}{\textit{Understanding and Generation}} \\[-1pt]

        Layer skipping
        & Candidate ratio 20\%
        & Candidate ratio 25\%;\newline
          $\delta_{\mathrm{LS}} = 0.0040$ \\

        FFN pruning
        & Prune ratio 20\%
        & Candidate prune ratio 20\%;\newline
          $\delta_{\mathrm{FFN}} = 0.150$ \\
          \hline

        \addlinespace[3pt]

        \rowcolor{gray!12}
        \multicolumn{3}{@{}l}{\textit{Generation only}} \\[-1pt]

        Diffusion-head cache
        & Fixed horizon 5
        & Candidate horizons 5/8/10;\newline
          $\delta_{\mathrm{DC}}^{\mathrm{mid}} = 0.35$ and
          $\delta_{\mathrm{DC}}^{\mathrm{high}} = 0.50$ \\

        Early exit
        & Disabled
& Activated in the final 10\% of the trajectory;
$\delta_{\mathrm{EE}}=0.42$ \\

        \bottomrule
    \end{tabularx}

    \caption{Mapping between FlashU and UCS operating points on Show-o2.}
    \label{tab:app_ucs_showo}
\end{table*}

\subsubsection{UCS on BAGEL}

Table~\ref{tab:app_ucs_bagel} gives the corresponding configuration on BAGEL. On the understanding side, the 80\% keep ratio of CE-Router corresponds to token-level compaction. On the generation side, routing-guided cache reuse and late-stage early exit are additionally enabled.

\begin{table*}[!ht]
    \centering
    \scriptsize
    \setlength{\tabcolsep}{2.4pt}
    \renewcommand{\arraystretch}{1.10}

    \begin{tabularx}{\textwidth}{
        @{}
        >{\raggedright\arraybackslash}p{0.27\textwidth}
        >{\raggedright\arraybackslash}p{0.23\textwidth}
        >{\raggedright\arraybackslash}X
        @{}
    }
        \toprule
        \textbf{Operator}
        & \textbf{FlashU}
        & \textbf{CE-Router + UCS} \\
        \midrule

        \rowcolor{gray!12}
        \multicolumn{3}{@{}l@{}}{\textit{Understanding}} \\[-1pt]

        Token compaction
        & Prune ratio 15\%
        & Prune ratio 20\% \\

        \addlinespace[1.5pt]

        Layer skipping
        & Skip ratio 10\%
        & Skip ratio 15\%;\newline
          $\delta_{\mathrm{LS}} = 0.0040$ \\

        \addlinespace[1.5pt]

        FFN pruning
        & Prune ratio 12\%
        & Candidate prune ratio 18\%;\newline
          $\delta_{\mathrm{FFN}} = 0.150$ \\
          \hline

        \addlinespace[3pt]

        \rowcolor{gray!12}
        \multicolumn{3}{@{}l@{}}{\textit{Generation}} \\[-1pt]

        Token compaction
        & --
        & Prune ratio 20\% \\

        \addlinespace[1.5pt]

        Layer skipping
        & Skip ratio 8\%
        & Skip ratio 12\%;\newline
          $\delta_{\mathrm{LS}} = 0.0040$ \\

        \addlinespace[1.5pt]

        FFN pruning
        & Prune ratio 10\%
        & Candidate prune ratio 15\%;\newline
          $\delta_{\mathrm{FFN}} = 0.150$ \\

        \addlinespace[1.5pt]

        Diffusion-head cache
        & Disabled
        & Candidate horizons 5/8/10;\newline
          $\delta_{\mathrm{DC}}^{\mathrm{mid}} = 0.35$,\newline
          $\delta_{\mathrm{DC}}^{\mathrm{high}} = 0.50$ \\

        \addlinespace[1.5pt]

        Denoising-step early exit
        & Disabled
        & Activated in the final 10\% of the trajectory;\newline
          $\delta_{\mathrm{EE}} = 0.42$ \\

        \bottomrule
    \end{tabularx}

    \caption{Mapping between FlashU and UCS operating points on BAGEL.}
    \label{tab:app_ucs_bagel}
\end{table*}

\begin{table*}[!ht]
    \centering
    \footnotesize
    \setlength{\tabcolsep}{3pt}
    \renewcommand{\arraystretch}{1.10}

    \begin{tabularx}{\textwidth}{
        @{}
        >{\raggedright\arraybackslash}p{0.31\textwidth}
        >{\raggedright\arraybackslash}X
        @{}
    }
        \toprule
        \textbf{Item}
        & \textbf{Setting} \\
        \midrule

        \rowcolor{gray!12}
        \multicolumn{2}{@{}l@{}}{\textit{Backbone}} \\[-1pt]

        Base model
        & Show-o2-1.5B \\

        Language backbone
        & Qwen2.5-1.5B-Instruct~\cite{yang2024qwen25} \\

        Vision encoder
        & SigLIP SO400M Patch14-384~\cite{zhai2023siglip} \\

        VAE
        & Wan2.1 VAE~\cite{wan2025} \\
        \hline

        \addlinespace[3pt]

        \rowcolor{gray!12}
        \multicolumn{2}{@{}l@{}}{\textit{Model Architecture}} \\[-1pt]

        Hidden size
        & 1,536 \\

        Routed layers
        & Last 12 Qwen decoder layers \\

        Maximum sequence length
        & 1,024 \\

        Latent grid
        & $27 \times 27 = 729$ \\

        Time embedding
        & Enabled; image-token slots are extended from 729 to 730 \\ \hline

        \addlinespace[3pt]

        \rowcolor{gray!12}
        \multicolumn{2}{@{}l@{}}{\textit{Router Configuration}} \\[-1pt]

        Router form
        & \texttt{Linear(hidden\_size,1)} + sigmoid + hard top-$K$ \\

        Router assignment
        & Independent routers for Und. and Gen. \\ \hline

        \addlinespace[3pt]

        \rowcolor{gray!12}
        \multicolumn{2}{@{}l@{}}{\textit{Input Configuration}} \\[-1pt]

        Training/inference resolution
        & $432 \times 432$ \\

        \bottomrule
    \end{tabularx}

    \caption{Backbone and router configuration of Show-o2 + UniMoD.}
    \label{tab:app_unimod_showo_arch}
\end{table*}

\begin{table*}[!t]
    \centering
    \scriptsize
    \setlength{\tabcolsep}{3pt}
    \renewcommand{\arraystretch}{1.10}

    \begin{tabularx}{\textwidth}{
        @{}
        >{\raggedright\arraybackslash}p{0.42\textwidth}
        >{\raggedright\arraybackslash}X
        @{}
    }
        \toprule
        \textbf{Parameter}
        & \textbf{Setting} \\
        \midrule

        \rowcolor{gray!12}
        \multicolumn{2}{@{}l@{}}{\textit{Batch and Data}} \\[-1pt]

        Per-GPU / global batch size
        & 8 / 64 \\

        Gradient accumulation
        & 1 \\

        Gen. data
        & 10K, aligned with Show-o2~\cite{xie2026showo2} \\
    
        Und. data
        & 10K, aligned with Show-o2~\cite{xie2026showo2} \\ \hline

        \addlinespace[3pt]

        \rowcolor{gray!12}
        \multicolumn{2}{@{}l@{}}{\textit{Optimization}} \\[-1pt]

        Training steps
        & 1,920 \\

        Optimizer
        & AdamW \\

        $\beta_1,\beta_2$ / $\epsilon$
        & 0.9, 0.999 / $10^{-8}$ \\

        Weight decay / max gradient norm
        & 0 / 1.0 \\

        Schedule / warmup
        & Cosine / 0.03 (57 steps) \\ \hline

        \addlinespace[3pt]

        \rowcolor{gray!12}
        \multicolumn{2}{@{}l@{}}{\textit{Learning Rates}} \\[-1pt]

        Und. vision path
        & $2 \times 10^{-6}$ \\

        Fusion projection
        & $1 \times 10^{-5}$ \\

        Other parameters
        & $1 \times 10^{-5}$ \\ \hline

        \addlinespace[3pt]

        \rowcolor{gray!12}
        \multicolumn{2}{@{}l@{}}{\textit{Objective and Other Settings}} \\[-1pt]

        Training loss
        & $\mathcal{L}_{\mathrm{NTP}}
           + 10\mathcal{L}_{\mathrm{flow}}$ \\

        Conditional dropout / \texttt{und\_max\_t0}
        & 0.0 / 1.0 \\

        Random seed
        & 10086 \\

        \bottomrule
    \end{tabularx}

    \caption{Training configuration of Show-o2 + UniMoD.}
    \label{tab:app_unimod_showo_training}
\end{table*}

\begin{table*}[!t]
    \centering

    \footnotesize
    \setlength{\tabcolsep}{4pt}
    \renewcommand{\arraystretch}{1.15}

    \begin{tabular*}{\textwidth}{
        @{\extracolsep{\fill}}
        l
        c
        c
        c
        c
        c
        @{}
    }
        \toprule

        \textbf{Benchmark}
        & \textbf{Batch Size}
        & \textbf{CFG Scale}
        & \textbf{Steps}
        & \textbf{Resolution}
        & \textbf{Devices} \\
        \midrule

        GenEval
        & 4
        & 7.5
        & 100
        & $432 \times 432$
        & 8 \\

        DPG-Bench
        & 4
        & 10
        & 50
        & $432 \times 432$
        & 8 \\

        \bottomrule
    \end{tabular*}

    \caption{Generation evaluation configuration of Show-o2 + UniMoD.}
    \label{tab:app_unimod_showo_eval}
\end{table*}

\subsubsection{One-Time Threshold Profiling}

The UCS thresholds are selected through one-time profiling. For each evaluation dataset, we sample 20 examples and run one actual inference pass, while recording the layer-skipping redundancy gap $g_{l,t}$, retained-token saliency leakage $\lambda_{l,t}$, router-cache agreement $\mathrm{RDA}_{t,t_{\mathrm{ref}}}$, and early-exit router-mask Jaccard similarity $J_{t,t-1}$ for the applicable operators.

We then replay multiple candidate thresholds offline without regenerating images and count the operations admitted by each configuration, including the average number of skipped layers, pruned-FFN admission rate, cache-extension rate, early-exit trigger rate, and average number of skipped denoising steps. The complete profiling procedure takes approximately 14.6 minutes, of which the generation portion takes approximately 9.5 minutes. This process is used only to quickly screen candidate operating points that balance accuracy risk and speed; the final configuration is still confirmed through one complete end-to-end inference evaluation.

\subsection{UniMoD Reproduction Details}
\label{sec:app_unimod}

The official UniMoD~\cite{mao2025unimod} code has not been released. Therefore, the results on Show-o2 and BAGEL are obtained from our implementation based on the paper description. The main paper reports only the results; this section provides the architecture, training, and evaluation settings required for reproduction.

\subsubsection{UniMoD on Show-o2}

Table~\ref{tab:app_unimod_showo_arch} summarizes the backbone and router configuration on Show-o2. Understanding and generation use independent routers, each performing top-$K$ token routing for its corresponding task.

Table~\ref{tab:app_unimod_showo_training} further lists the training settings of the understanding router on Show-o2. The generation router is initialized independently from the original Show-o2 checkpoint, trained for 1,920 steps using only 10K generation samples, and uses the same optimizer, scheduler, precision, and distributed settings.

Table~\ref{tab:app_unimod_showo_eval} gives the inference settings of this reproduction on the two generation benchmarks. For DPG-Bench, four images are generated for each prompt.

\subsubsection{UniMoD on BAGEL}

The BAGEL router consists of LayerNorm, a two-layer MLP, hard top-$K$, and a straight-through gate. The understanding router operates on ViT tokens, whereas the generation router operates on VAE latent tokens; text and structure tokens are always retained. Table~\ref{tab:app_unimod_bagel} summarizes the training configurations of the two routers.

\begin{table*}[!ht]
    \centering
    \footnotesize
    \setlength{\tabcolsep}{2.6pt}
    \renewcommand{\arraystretch}{1.10}

    \begin{tabularx}{\textwidth}{
        @{}
        >{\raggedright\arraybackslash}p{0.36\textwidth}
        >{\raggedright\arraybackslash}p{0.29\textwidth}
        >{\raggedright\arraybackslash}X
        @{}
    }
        \toprule
        \textbf{Setting}
        & \textbf{Und. Router}
        & \textbf{Gen. Router} \\
        \midrule

        \rowcolor{gray!12}
        \multicolumn{3}{@{}l@{}}{\textit{Routing Configuration}} \\[-1pt]

        Routed tokens
        & ViT tokens
        & VAE latent tokens \\

        Always-retained tokens
        & Text and structure tokens
        & Text and structure tokens \\

        Routed layers
        & Last 12 layers
        & Last 12 layers \\

        Router bottleneck
        & Hidden size / 4
        & Hidden size / 4 \\
        \hline

        \addlinespace[3pt]

        \rowcolor{gray!12}
        \multicolumn{3}{@{}l@{}}{\textit{Training Configuration}} \\[-1pt]

        Training steps
        & 4K
        & 4K \\

        Keep ratio
        & 0.8
        & 0.95 \\

        Learning rate
        & $3 \times 10^{-5}$
        & $3 \times 10^{-5}$ \\

        Consistency weight
        & 0.5
        & 0.5 \\ \hline

        \addlinespace[3pt]

        \rowcolor{gray!12}
        \multicolumn{3}{@{}l@{}}{\textit{Data and Input}} \\[-1pt]

        Training data
        & 10K Und.
        & 100K Gen. \\

        Training/inference resolution
        & $432 \times 432$
        & $432 \times 432$ \\

        \bottomrule
    \end{tabularx}

    \caption{Router and training configuration of BAGEL + UniMoD.}
    \label{tab:app_unimod_bagel}
\end{table*}


\section{Implementation and Statistical Details for Figure~1}
\label{sec:app_token_importance}

This section provides the implementation details and statistical protocols for the token-importance analyses in Figure~1 of the main paper. All experiments are conducted on Show-o2-1.5B. Figure~1(a) directly analyzes the uncompressed dense model. For Figures~1(b) and~1(c), we freeze the original backbone and train two task-specific probes with the same architecture but independent parameters. The main paper presents the primary definitions and findings; below, we provide only the details required for reproduction.

\subsection{Received-Attention Analysis}
\label{sec:app_received_attention}

Figure~1(a) uses self-attention from the dense model as a proxy for token importance on the understanding side. For an input sample $x$, let
\begin{equation}
    A_l(x)\in\mathbb{R}^{N_x\times N_x}
\end{equation}
denote the attention map at layer $l$ after averaging over attention heads, where $A_l(x)_{j,i}$ is the attention assigned by query token $j$ to token $i$. The mean attention received by token $i$ at layer $l$ is defined as
\begin{equation}
    r_{l,i}(x)
    =
    \frac{1}{N_x}
    \sum_{j=1}^{N_x} A_l(x)_{j,i}.
\end{equation}
This score is used only as a proxy for routing importance and is not equivalent to a router score learned with the task objective.

In practice, we extract attention weights from the decoder layers and first average them over the head dimension. For inputs of different lengths, a valid-token mask excludes padded queries and keys. Figure~1(a) reports the average over 300 randomly sampled understanding examples.

To improve the visibility of within-layer variation, we first apply a $\log(1+x)$ transformation to the sample-averaged scores at each layer, then clip them to the layer-wise 1st and 99th percentiles and normalize the resulting values to $[0,1]$. Therefore, the colors in Figure~1(a) primarily represent the relative importance of token positions within the same layer and are not intended for direct comparison of absolute attention mass across layers.

\subsection{Training Task-Specific Probes}
\label{sec:app_probe_training}

Figures~1(b) and~1(c) use one understanding probe and one generation probe. Both are built on the same Show-o2-1.5B baseline and use the same token-wise scoring architecture, but their parameters are independent. They are used only to analyze task-specific token-importance patterns and do not adopt the cross-task Core-Expansion parameter sharing of CE-Router.

During training, the original Show-o2 backbone is frozen, and only the probe router and LoRA adapter are updated. Table~\ref{tab:app_probe_training} summarizes the training configurations of the two probes.

\begin{table*}[!t]
    \centering
    \footnotesize
    \setlength{\tabcolsep}{4.5pt}
    \renewcommand{\arraystretch}{1.08}
    \begin{tabularx}{\linewidth}{lXX}
        \toprule
        \textbf{Setting} & \textbf{Understanding Probe} & \textbf{Generation Probe} \\
        \midrule
        Backbone & Show-o2-1.5B & Show-o2-1.5B \\
    Training data & 10K, aligned with Show-o2~\cite{xie2026showo2} & 10K, aligned with Show-o2~\cite{xie2026showo2} \\
        Training steps & 1,920 & 1,920 \\
        Trainable components & Probe router + LoRA & Probe router + LoRA \\
        LoRA rank & 16 & 16 \\
        Task objective & VQA next-token prediction & T2I flow matching \\
        Consistency weight $\lambda_{\mathrm{con}}$ & 0.25 & 0.25 \\
        \bottomrule
    \end{tabularx}
    \caption{Training configurations of the task-specific probes.}
    \label{tab:app_probe_training}
\end{table*}

Both probes use the task loss and dense-model consistency loss defined in the main paper. The consistency target uses the final hidden states from routed and dense execution immediately before the corresponding task head, and the mean-squared error is computed directly over the complete $[\text{batch},\text{seqlen},\text{hidden}]$ tensor. This training setup makes the probe score indicate which tokens a routing policy tends to prioritize when optimized for only one task.

\subsection{Probe Scores and Top-$k$ Statistics}
\label{sec:app_probe_statistics}

After training, we directly extract the token-wise logits produced by the task-specific probe at each backbone layer, rank them independently within that layer, and select the top-$k$ tokens.

Because the sigmoid function is monotonic, ranking by router logits or sigmoid scores produces the same top-$k$ set. All set-similarity statistics are first computed independently at each layer and then averaged over samples and layers.

\subsection{Timestep Variation During Generation}
\label{sec:app_timestep_variation}

Figure~1(b) uses the Jaccard distance defined in the main paper to measure how the important-token set selected by the generation probe changes over the denoising process. For each generation sample, we compare the top-$k$ set at the current timestep with that at the initial evaluated timestep $t_0$ independently at every analyzed layer, and then average over samples and layers.

Each point in the figure therefore represents the average set difference between the corresponding generation step and $t_0$. A larger Jaccard distance indicates a more pronounced change in the ranking and selection of important tokens relative to the beginning of generation.

\subsection{Cross-Task Probe Transfer}
\label{sec:app_probe_transfer}

Figure~1(c) compares how the two probes transfer across task inputs.

For Understanding probe on generation samples, we run both probes on the generation input at each evaluated timestep, compute the directional top-$k$ coverage defined in the main paper independently at every layer, and then average over samples and layers. The blue horizontal bars in Figure~1(c) report these results, which vary with the generation step.

For Generation probe on understanding samples, we run both probes on understanding inputs and compute the average top-$k$ coverage in the same manner. Because understanding has no denoising trajectory, this setting produces a single aggregate result, shown by the pink vertical dashed line at 84.45\%.

Figure~1(c) therefore compares the transfer performance of the understanding probe across different generation steps with the single aggregate transfer result of the generation probe on understanding samples.


\begin{table*}[h]
    \centering
    
    \small
    \setlength{\tabcolsep}{3.6pt}
    \renewcommand{\arraystretch}{1.08}

    \resizebox{\linewidth}{!}{%
    \begin{tabular}{c|l|cccccccc|cc}
        \toprule
        \textbf{Scale}
        & \textbf{Method}
        & \textbf{AI2D} $\uparrow$
        & \textbf{GQA} $\uparrow$
        & \textbf{MMBench} $\uparrow$
        & \textbf{MME} $\uparrow$
        & \textbf{MMMU} $\uparrow$
        & \textbf{MMStar} $\uparrow$
        & \textbf{SEEDBench} $\uparrow$
        & \textbf{Rel. Avg. (\%)} $\uparrow$
        & \textbf{TFLOPs} $\downarrow$
        & \textbf{Latency (ms)} $\downarrow$ \\
        \midrule

        \multirow{5}{*}{\textbf{1.5B}}
        & Show-o2
        & 68.98
        & 59.95
        & 67.35
        & 1753.44
        & 36.56
        & 43.39
        & 65.63
        & 100.00
        & 6.37
        & 474.41 \\

        & Show-o2 + FlashU
        & 52.46
        & 48.31
        & 48.80
        & 1343.54
        & 33.00
        & 39.62
        & 62.40
        & 83.19
        & 6.27
        & 483.85 \\

        & Show-o2 + UniMoD
        & 65.71
        & 57.16
        & 63.06
        & 1603.07
        & 35.20
        & 41.18
        & 64.62
        & 95.04
        & 5.42
        & 461.52 \\

        & \cellcolor{oursrow}Show-o2 + CE-Router (Ours)
        & \cellcolor{oursrow}\textbf{67.75}
        & \cellcolor{oursrow}\textbf{59.88}
        & \cellcolor{oursrow}\textbf{67.36}
        & \cellcolor{oursrow}\textbf{1717.29}
        & \cellcolor{oursrow}\underline{36.22}
        & \cellcolor{oursrow}\textbf{43.33}
        & \cellcolor{oursrow}\textbf{65.99}
        & \cellcolor{oursrow}\textbf{99.36}
        & \cellcolor{oursrow}\underline{3.04}
        & \cellcolor{oursrow}\underline{300.32} \\

        & \cellcolor{oursrow}Show-o2 + CE-Router + UCS
        & \cellcolor{oursrow}\underline{67.23}
        & \cellcolor{oursrow}\underline{57.28}
        & \cellcolor{oursrow}\underline{66.84}
        & \cellcolor{oursrow}\underline{1687.10}
        & \cellcolor{oursrow}\textbf{36.55}
        & \cellcolor{oursrow}\underline{42.57}
        & \cellcolor{oursrow}\underline{65.39}
        & \cellcolor{oursrow}\underline{98.03}
        & \cellcolor{oursrow}\textbf{2.73}
        & \cellcolor{oursrow}\textbf{245.25} \\

        \midrule

        \multirow{5}{*}{\textbf{7B}}
        & BAGEL
        & 87.76
        & 66.36
        & 84.79
        & 2349.85
        & 51.33
        & 69.49
        & 72.86
        & 100.00
        & 25.32
        & 487.68 \\

        & BAGEL + FlashU
        & 81.77
        & 62.22
        & 81.93
        & 2128.93
        & 49.17
        & 62.71
        & 72.08
        & 94.16
        & \underline{22.47}
        & 465.48 \\

        & BAGEL + UniMoD
        & 85.27
        & \underline{63.47}
        & 82.64
        & \underline{2168.56}
        & 49.89
        & 66.14
        & 72.22
        & 96.29
        & 23.58
        & 458.69 \\

        & \cellcolor{oursrow}BAGEL + CE-Router (Ours)
        & \cellcolor{oursrow}\textbf{87.95}
        & \cellcolor{oursrow}\textbf{66.12}
        & \cellcolor{oursrow}\textbf{84.79}
        & \cellcolor{oursrow}\textbf{2339.69}
        & \cellcolor{oursrow}\textbf{51.78}
        & \cellcolor{oursrow}\textbf{68.96}
        & \cellcolor{oursrow}\textbf{72.99}
        & \cellcolor{oursrow}\textbf{99.96}
        & \cellcolor{oursrow}23.31
        & \cellcolor{oursrow}\underline{417.76} \\

        & \cellcolor{oursrow}BAGEL + CE-Router + UCS
        & \cellcolor{oursrow}\underline{85.92}
        & \cellcolor{oursrow}62.80
        & \cellcolor{oursrow}\underline{83.11}
        & \cellcolor{oursrow}2140.82
        & \cellcolor{oursrow}\underline{50.46}
        & \cellcolor{oursrow}\underline{66.33}
        & \cellcolor{oursrow}\underline{72.62}
        & \cellcolor{oursrow}\underline{96.44}
        & \cellcolor{oursrow}\textbf{17.27}
        & \cellcolor{oursrow}\textbf{403.34} \\

        \bottomrule
    \end{tabular}%
    }
    \caption{
        Comparison of multimodal understanding performance and inference efficiency.
        \textbf{Bold} and \underline{underlined} values denote the best and
        second-best results.
    }
    \label{tab:understanding_results}

\end{table*}

\definecolor{oursrow}{HTML}{F2F2F2}
\begin{table*}[!t]
    \centering

    \small
    \setlength{\tabcolsep}{3.2pt}
    \renewcommand{\arraystretch}{1.08}

    \resizebox{\linewidth}{!}{%
    \begin{tabular}{c|l|cccccccc|c}
        \toprule
        \textbf{Scale}
        & \textbf{Method}
        & \textbf{Corr. Img.} $\uparrow$
        & \textbf{Corr. Prompt} $\uparrow$
        & \textbf{Single Obj.} $\uparrow$
        & \textbf{Two Obj.} $\uparrow$
        & \textbf{Counting} $\uparrow$
        & \textbf{Colors} $\uparrow$
        & \textbf{Position} $\uparrow$
        & \textbf{Color Attr.} $\uparrow$
        & \textbf{Overall} $\uparrow$ \\
        \midrule

        \multirow{5}{*}{\textbf{1.5B}}
        & Show-o2
        & 70.61
        & 82.64
        & 98.44
        & 85.10
        & 57.81
        & 86.70
        & 41.75
        & 58.00
        & 71.30 \\

        & Show-o2 + FlashU
        & 63.38
        & 77.76
        & \underline{98.44}
        & 72.47
        & \underline{45.00}
        & \textbf{83.78}
        & 31.50
        & \underline{53.75}
        & 64.16 \\

        & Show-o2 + UniMoD
        & 62.21
        & \textbf{79.75}
        & \textbf{98.75}
        & 72.73
        & 37.81
        & 81.91
        & 33.75
        & 52.00
        & 62.83 \\

        & \cellcolor{oursrow}Show-o2 + CE-Router (Ours)
        & \cellcolor{oursrow}\textbf{65.60}
        & \cellcolor{oursrow}\underline{79.20}
        & \cellcolor{oursrow}\underline{98.44}
        & \cellcolor{oursrow}\textbf{75.51}
        & \cellcolor{oursrow}\textbf{49.06}
        & \cellcolor{oursrow}82.45
        & \cellcolor{oursrow}\textbf{40.25}
        & \cellcolor{oursrow}52.25
        & \cellcolor{oursrow}\textbf{66.33} \\

        & \cellcolor{oursrow}Show-o2 + CE-Router + UCS
        & \cellcolor{oursrow}\underline{63.83}
        & \cellcolor{oursrow}\textbf{79.75}
        & \cellcolor{oursrow}97.19
        & \cellcolor{oursrow}\underline{74.49}
        & \cellcolor{oursrow}42.50
        & \cellcolor{oursrow}\underline{83.51}
        & \cellcolor{oursrow}\underline{34.50}
        & \cellcolor{oursrow}\textbf{54.50}
        & \cellcolor{oursrow}\underline{64.45} \\

        \midrule

        \multirow{5}{*}{\textbf{7B}}
        & BAGEL
        & 78.16
        & 89.33
        & 98.44
        & 87.88
        & 73.75
        & 87.23
        & 62.50
        & 63.00
        & 78.80 \\

        & BAGEL + FlashU
        & 71.52
        & \textbf{91.32}
        & \textbf{98.75}
        & \underline{86.36}
        & \textbf{70.00}
        & 79.52
        & 46.50
        & 53.75
        & 72.48 \\

        & BAGEL + UniMoD
        & 53.25
        & 80.47
        & 88.44
        & 57.58
        & 39.38
        & 72.34
        & 32.75
        & 34.50
        & 54.16 \\

        & \cellcolor{oursrow}BAGEL + CE-Router (Ours)
        & \cellcolor{oursrow}\textbf{75.45}
        & \cellcolor{oursrow}\underline{86.80}
        & \cellcolor{oursrow}\textbf{98.75}
        & \cellcolor{oursrow}\textbf{88.13}
        & \cellcolor{oursrow}\underline{60.00}
        & \cellcolor{oursrow}\textbf{94.15}
        & \cellcolor{oursrow}\textbf{52.25}
        & \cellcolor{oursrow}\textbf{62.25}
        & \cellcolor{oursrow}\textbf{75.92} \\

        & \cellcolor{oursrow}BAGEL + CE-Router + UCS
        & \cellcolor{oursrow}\underline{72.65}
        & \cellcolor{oursrow}84.81
        & \cellcolor{oursrow}\underline{97.50}
        & \cellcolor{oursrow}85.10
        & \cellcolor{oursrow}56.56
        & \cellcolor{oursrow}\underline{91.22}
        & \cellcolor{oursrow}\underline{49.25}
        & \cellcolor{oursrow}\underline{59.25}
        & \cellcolor{oursrow}\underline{73.15} \\

        \bottomrule
    \end{tabular}%
    }
    \caption{
            Category-wise comparison of compositional image generation performance on GenEval.
        \textbf{Bold} and \underline{underlined} values denote the best and
        second-best results.
    }
    \label{tab:geneval_detailed}
\end{table*}

\definecolor{oursrow}{HTML}{F2F2F2}

\begin{table*}[!t]
    \centering

    \scriptsize
    \setlength{\tabcolsep}{1.8pt}
    \renewcommand{\arraystretch}{1.08}

    \resizebox{\linewidth}{!}{%
    \begin{tabular}{c|l|*{18}{c}|c}
        \toprule
        \textbf{Scale}
        & \textbf{Method}
        & \textbf{L1 Rel.} $\uparrow$
        & \textbf{L1 Glob.} $\uparrow$
        & \textbf{L1 Attr.} $\uparrow$
        & \textbf{L1 Ent.} $\uparrow$
        & \textbf{L1 Other} $\uparrow$
        & \textbf{L2 Attr.-Col.} $\uparrow$
        & \textbf{L2 Attr.-Oth.} $\uparrow$
        & \textbf{L2 Attr.-Shp.} $\uparrow$
        & \textbf{L2 Attr.-Size} $\uparrow$
        & \textbf{L2 Attr.-Tex.} $\uparrow$
        & \textbf{L2 Ent.-Part} $\uparrow$
        & \textbf{L2 Ent.-State} $\uparrow$
        & \textbf{L2 Ent.-Whole} $\uparrow$
        & \textbf{L2 Glob.} $\uparrow$
        & \textbf{L2 Oth.-Cnt.} $\uparrow$
        & \textbf{L2 Oth.-Text} $\uparrow$
        & \textbf{L2 Rel.-NSp.} $\uparrow$
        & \textbf{L2 Rel.-Sp.} $\uparrow$
        & \textbf{Overall} $\uparrow$ \\
        \midrule

        \multirow{5}{*}{\textbf{1.5B}}
        & Show-o2
        & 89.86
        & 86.32
        & 91.24
        & 91.47
        & 90.57
        & 92.02
        & 88.72
        & 89.41
        & 82.19
        & 93.25
        & 87.53
        & 87.32
        & 92.73
        & 86.32
        & 90.45
        & 91.03
        & 88.75
        & 89.93
        & 85.25 \\

        & Show-o2 + FlashU
        & 92.34
        & 79.64
        & 87.77
        & 87.72
        & \underline{70.40}
        & 91.07
        & 84.59
        & 77.73
        & 70.66
        & 89.41
        & \underline{84.77}
        & \underline{83.21}
        & 88.95
        & 79.64
        & 68.50
        & \textbf{78.00}
        & 85.53
        & 92.79
        & 81.06 \\

        & Show-o2 + UniMoD
        & 92.61
        & 80.24
        & \underline{89.04}
        & 88.22
        & 69.20
        & \underline{92.57}
        & 84.92
        & 79.48
        & \textbf{73.97}
        & \textbf{90.56}
        & 83.79
        & \underline{83.21}
        & 89.71
        & 80.24
        & 69.00
        & 70.00
        & \underline{86.79}
        & 92.99
        & 81.72 \\

        & \cellcolor{oursrow}Show-o2 + CE-Router (Ours)
        & \cellcolor{oursrow}\textbf{92.88}
        & \cellcolor{oursrow}\underline{81.76}
        & \cellcolor{oursrow}\textbf{89.22}
        & \cellcolor{oursrow}\textbf{88.98}
        & \cellcolor{oursrow}\textbf{72.40}
        & \cellcolor{oursrow}\textbf{92.62}
        & \cellcolor{oursrow}\textbf{86.25}
        & \cellcolor{oursrow}\underline{79.91}
        & \cellcolor{oursrow}\underline{73.55}
        & \cellcolor{oursrow}\underline{90.32}
        & \cellcolor{oursrow}\textbf{86.91}
        & \cellcolor{oursrow}\textbf{85.11}
        & \cellcolor{oursrow}\textbf{89.98}
        & \cellcolor{oursrow}\underline{81.76}
        & \cellcolor{oursrow}\textbf{72.00}
        & \cellcolor{oursrow}\underline{74.00}
        & \cellcolor{oursrow}\textbf{87.42}
        & \cellcolor{oursrow}\textbf{93.24}
        & \cellcolor{oursrow}\textbf{82.86} \\

        & \cellcolor{oursrow}Show-o2 + CE-Router + UCS
        & \cellcolor{oursrow}\underline{92.73}
        & \cellcolor{oursrow}\textbf{82.37}
        & \cellcolor{oursrow}88.66
        & \cellcolor{oursrow}\underline{88.27}
        & \cellcolor{oursrow}70.00
        & \cellcolor{oursrow}92.17
        & \cellcolor{oursrow}\underline{85.81}
        & \cellcolor{oursrow}\textbf{80.35}
        & \cellcolor{oursrow}71.07
        & \cellcolor{oursrow}89.72
        & \cellcolor{oursrow}84.38
        & \cellcolor{oursrow}82.79
        & \cellcolor{oursrow}\underline{89.79}
        & \cellcolor{oursrow}\textbf{82.37}
        & \cellcolor{oursrow}\underline{69.50}
        & \cellcolor{oursrow}72.00
        & \cellcolor{oursrow}\underline{86.79}
        & \cellcolor{oursrow}\underline{93.12}
        & \cellcolor{oursrow}\underline{81.85} \\

        \midrule

        \multirow{5}{*}{\textbf{7B}}
        & BAGEL
        & 93.69
        & 83.58
        & 88.22
        & 89.71
        & 84.80
        & 92.72
        & 85.80
        & 79.03
        & 69.83
        & 88.08
        & 82.22
        & 83.94
        & 91.67
        & 83.58
        & 83.00
        & 92.00
        & 88.05
        & 94.06
        & 83.95 \\

        & BAGEL + FlashU
        & 92.38
        & \underline{80.24}
        & \underline{87.33}
        & 87.12
        & 69.60
        & \underline{91.22}
        & 82.82
        & 80.35
        & \underline{69.42}
        & \underline{88.69}
        & 84.96
        & \textbf{83.42}
        & 88.10
        & \underline{80.24}
        & 69.00
        & 72.00
        & 85.53
        & 92.83
        & 79.59 \\

        & BAGEL + UniMoD
        & \underline{93.43}
        & 78.42
        & 83.74
        & \underline{88.87}
        & \underline{81.60}
        & 89.11
        & 79.26
        & 78.60
        & 66.94
        & 82.87
        & \textbf{87.89}
        & 81.73
        & \underline{90.40}
        & 78.42
        & \underline{79.00}
        & \textbf{92.00}
        & \underline{87.42}
        & \underline{93.82}
        & 80.72 \\

        & \cellcolor{oursrow}BAGEL + CE-Router (Ours)
        & \cellcolor{oursrow}\textbf{94.27}
        & \cellcolor{oursrow}79.94
        & \cellcolor{oursrow}86.31
        & \cellcolor{oursrow}\textbf{90.16}
        & \cellcolor{oursrow}\textbf{84.40}
        & \cellcolor{oursrow}90.37
        & \cellcolor{oursrow}\underline{83.59}
        & \cellcolor{oursrow}\textbf{81.65}
        & \cellcolor{oursrow}69.01
        & \cellcolor{oursrow}86.09
        & \cellcolor{oursrow}\underline{87.70}
        & \cellcolor{oursrow}82.15
        & \cellcolor{oursrow}\textbf{92.03}
        & \cellcolor{oursrow}79.94
        & \cellcolor{oursrow}\textbf{84.00}
        & \cellcolor{oursrow}\underline{86.00}
        & \cellcolor{oursrow}\textbf{89.31}
        & \cellcolor{oursrow}\textbf{94.60}
        & \cellcolor{oursrow}\textbf{83.49} \\

        & \cellcolor{oursrow}BAGEL + CE-Router + UCS
        & \cellcolor{oursrow}92.34
        & \cellcolor{oursrow}\textbf{80.55}
        & \cellcolor{oursrow}\textbf{88.26}
        & \cellcolor{oursrow}87.51
        & \cellcolor{oursrow}66.40
        & \cellcolor{oursrow}\textbf{91.52}
        & \cellcolor{oursrow}\textbf{84.92}
        & \cellcolor{oursrow}\underline{81.22}
        & \cellcolor{oursrow}\textbf{69.83}
        & \cellcolor{oursrow}\textbf{89.84}
        & \cellcolor{oursrow}84.18
        & \cellcolor{oursrow}\underline{83.21}
        & \cellcolor{oursrow}88.73
        & \cellcolor{oursrow}\textbf{80.55}
        & \cellcolor{oursrow}65.00
        & \cellcolor{oursrow}72.00
        & \cellcolor{oursrow}83.65
        & \cellcolor{oursrow}92.91
        & \cellcolor{oursrow}\underline{80.77} \\

        \bottomrule
    \end{tabular}%
    }
    \caption{
        Category-wise comparison of text-to-image prompt-following performance on DPG-Bench.
        \textbf{Bold} and \underline{underlined} values denote the best and
        second-best results.
    }
    \label{tab:dpgbench_detailed}
\end{table*}

\section{Detailed Experimental Results}
\label{sec:app_detailed_results}

Tables~\ref{tab:understanding_results}, \ref{tab:geneval_detailed}, and~\ref{tab:dpgbench_detailed} provide the benchmark- and category-level results underlying the aggregate comparisons in the main paper.

\textbf{Multimodal Understanding.}
Table~\ref{tab:understanding_results} reports detailed results on the seven understanding benchmarks. CE-Router consistently preserves the dense-model performance across both backbones, retaining 99.36\% and 99.96\% of the original average scores on Show-o2 and BAGEL, respectively, and outperforming the other accelerated methods in relative average performance. After applying UCS, the resulting models further reduce TFLOPs and latency while still maintaining the strongest overall understanding performance among the accelerated baselines.

\textbf{Text-to-Image Generation.}
Tables~\ref{tab:geneval_detailed} and~\ref{tab:dpgbench_detailed} present the category-wise results on GenEval and DPG-Bench. CE-Router achieves the highest overall scores among the accelerated methods on both backbones and both benchmarks, with consistent advantages across object composition, attributes, spatial relations, and prompt-following categories. CE-Router with UCS largely retains these category-level gains and continues to outperform FlashU and UniMoD in overall generation performance, confirming that the additional runtime acceleration does not rely on a disproportionate degradation in any single evaluation dimension.

\section{Additional Qualitative Results}
\label{sec:app_additional_visuals}

\begin{figure*}[!t]
    \centering
    \includegraphics[width=\linewidth]{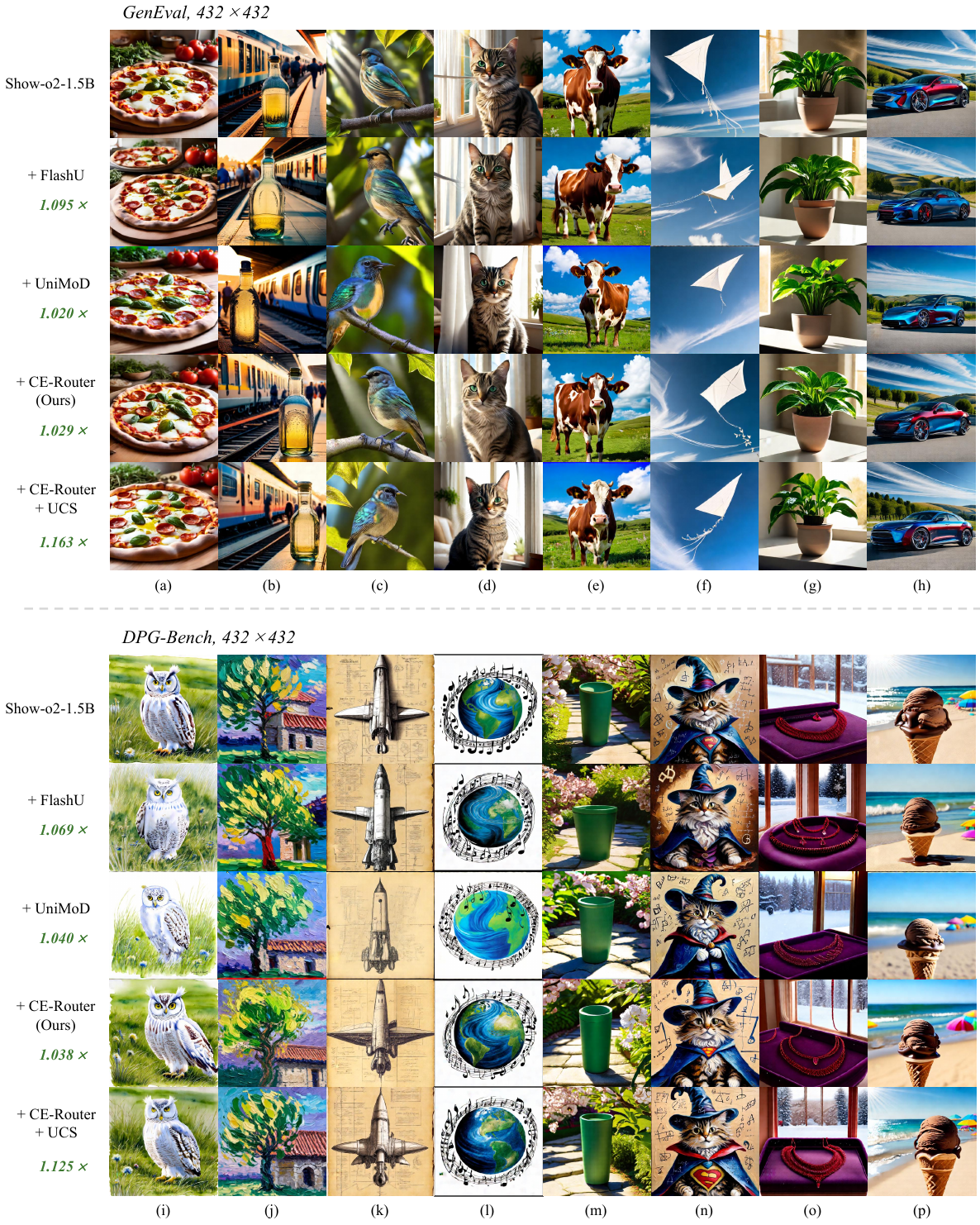}
    \caption{
        Additional qualitative comparisons on Show-o2-1.5B at
        $432\times432$ resolution.
        The upper and lower panels show examples from GenEval and
        DPG-Bench, respectively.
        Each column uses the same prompt across the dense backbone,
        FlashU, UniMoD, CE-Router, and CE-Router with UCS.
        The green values report end-to-end latency speedups relative to
        the dense Show-o2 backbone on the corresponding benchmark.
    }
    \label{fig:showo}
\end{figure*}

\begin{figure*}[!t]
    \centering
    \includegraphics[width=\linewidth]{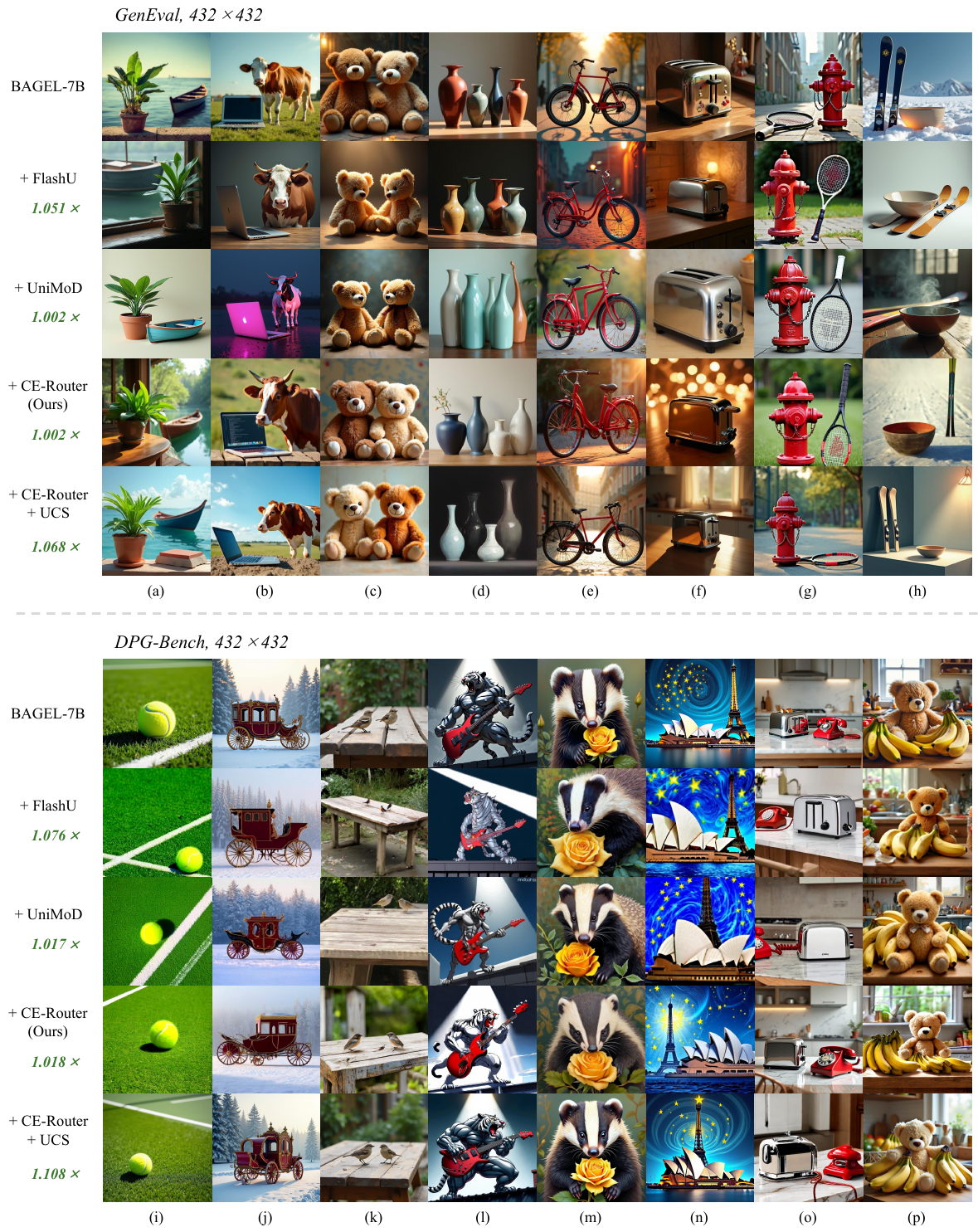}
    \caption{
        Additional qualitative comparisons on BAGEL-7B-MoT at
        $432\times432$ resolution.
        The upper and lower panels show examples from GenEval and
        DPG-Bench, respectively.
        Each column uses the same prompt across the dense backbone,
        FlashU, UniMoD, CE-Router, and CE-Router with UCS.
        The green values report end-to-end latency speedups relative to
        the dense BAGEL backbone on the corresponding benchmark.
    }
    \label{fig:bagel}
\end{figure*}

Figures~\ref{fig:showo} and~\ref{fig:bagel} provide additional qualitative comparisons on Show-o2-1.5B and BAGEL-7B-MoT, respectively. Across both backbones and benchmarks, CE-Router produces the most consistent visual quality among the accelerated methods. It better preserves object identity, geometry, fine-grained details, and spatial composition, particularly for prompts involving multiple objects, animals, vehicles, and small structural elements. In contrast, FlashU and UniMoD more frequently introduce malformed structures, missing or fused objects, attribute drift, and inconsistent relative placement. CE-Router with UCS largely retains these qualitative advantages while providing greater end-to-end acceleration, further demonstrating its favorable quality--efficiency trade-off.

\textbf{Prompts for Figure~\ref{fig:showo}.}
The prompts corresponding to columns (a)--(p) are listed below.

\begingroup
\small
\setlength{\parindent}{0pt}

\textbf{GenEval:}\par
\textbf{(a)} \textit{A photo of a pizza   A mouthwatering photo of a freshly baked pizza resting on a wooden board, showcasing its perfect golden-brown crust. The pizza is topped with bubbling melted mozzarella cheese, vibrant slices of pepperoni, fresh green basil leaves, and a drizzle of olive oil that glistens in the light. In the background, there\u2019s a warm, rustic kitchen setting with herbs, spices, and ripe tomatoes sitting nearby, adding to the atmosphere of authentic Italian cuisine. The image is shot in high definition with soft, natural lighting, highlighting the texture of the crust and the rich colors of the toppings.}\\

\textbf{(b)} \textit{A photo of a bottle right of a train.   A high-resolution photograph featuring a glass bottle placed to the right of a train. The scene is set outdoors at a busy railway station during golden hour, capturing warm lighting that reflects off the train's metallic surface and the transparent glass of the bottle. The bottle is slightly weathered, with intricate design patterns visible on its surface, while condensation beads glisten in the sunlight, indicating it contains a chilled liquid. The train, painted in vibrant colors, has a modern design with visible windows and doors. Behind the train, the station is alive with activity — blurred figures of passengers walking by, a bench in the background, and a faint haze in the air suggesting motion and life. The photo perfectly balances the contrast between the bottle's stillness and the energy of the station, showcasing fine details in the textures and lighting.}\\

\textbf{(c)} \textit{A photo of a bird.   A highly detailed and lifelike photo of a bird perched gracefully on a tree branch in a lush forest setting. The bird's feathers are vibrant and finely textured, showcasing a mix of shimmering blues, emerald greens, and soft golden accents. Sunlight filters through the canopy above, creating a beautiful play of light and shadows on the bird and the surrounding leaves. The background is slightly out of focus, emphasizing the radiant bird while still capturing a sense of depth and harmony in the natural environment.}\\

\textbf{(d)} \textit{A photo of a cat.   A highly detailed and realistic portrait of a tabby cat sitting near a sunny window, with light cascading through soft white curtains, illuminating its fur. The cat has striking emerald-green eyes that reflect the surrounding room, and its whiskers are long and delicate. Every strand of fur is rendered with photo-realistic texture, from the stripes on its head to the soft fluff on its chest. The background features a hint of cozy furniture and a few potted plants, adding warmth and depth to the scene.}\\

\textbf{(e)} \textit{A photo of a cow. A high-resolution photo of a healthy, serene cow standing in a lush green field under a bright blue sky dotted with fluffy white clouds. The cow's coat is smooth and shiny, showcasing a mix of earthy tones like deep brown with white patches. Its large, gentle eyes are illuminated by the sunlight, and its ears are perked up, seemingly aware of its surroundings. In the background, there are rolling hills and scattered wildflowers, contributing to the overall idyllic farm setting.}\\

\textbf{(f)} \textit{A photo of a white kite A high-resolution photograph of a strikingly elegant white kite gliding gracefully through the sky during a clear, sunny day. The kite appears to be made of lightweight material, with sharp, clean edges perfectly catching the light. The vibrant blue sky serves as a smooth background, dotted with wispy white clouds that highlight the kite's bright and airy design. The tail of the kite trails behind it in a mesmerizing rhythm, adorned with small bows or tassels fluttering gently in the breeze. The scene is serene and captures the beauty of simplicity and the joy of kite flying.}\\

\textbf{(g)} \textit{A photo of a potted plant. A crisp, high-resolution photo of a vibrant, lush potted plant placed on a clean, minimalistic tabletop. The plant has glossy green leaves that radiate health, with intricate vein patterns reflecting the soft, natural sunlight streaming in from a nearby window. The pot is modern and sleek, either ceramic with a matte finish or lightly textured terracotta, adding an earthy and contemporary vibe. The background features a neutral, slightly blurred indoor setting, creating a peaceful and inviting ambiance. Dust motes float subtly in the warm golden light, enhancing the serene, fresh, and natural aesthetic of the scene.}\\

\textbf{(h)} \textit{A photo of a car.   A high-resolution photograph capturing a sleek modern car, gleaming under natural daylight. The car features a stylish exterior design, with aerodynamic curves, sporty alloy wheels, and a glossy metallic finish in a vibrant hue, such as electric blue or ruby red. It is parked on an open road surrounded by a picturesque landscape of rolling hills, lined with lush green trees and a clear blue sky dotted with wispy white clouds. The scene is imbued with a sense of tranquility, while the car itself exudes luxury, performance, and innovation.}\par

\medskip
\textbf{DPG-Bench:}\par
\textbf{(i)} \textit{a detailed watercolor painting that captures a majestic snowy owl with its pristine white feathers standing in the midst of a lush green field. The owl's bright yellow eyes are a stark contrast to the soft hues of the grass, and its feathers are intricately detailed, giving a sense of texture to the artwork. The field is dotted with wildflowers and the occasional blade of grass that sways gently, suggesting a light breeze in this tranquil scene.}\\

\textbf{(j)} \textit{an impressionistic painting that features a vibrant array of colors, depicting a tree with a swirl of green and yellow leaves next to an old stone building with a red-tiled roof. The brush strokes are thick and visible, giving the painting a textured look, and the tree's branches seem to dance around the building's edges. The sky in the background is a mix of blues and purples, suggesting either dawn or dusk.}\\

\textbf{(k)} \textit{a detailed sketch of a space shuttle, rendered in the intricate, technical style reminiscent of Leonardo da Vinci's famous drawings. The shuttle is depicted with numerous annotations and measurements, showcasing its complex design and structure. The paper on which it is drawn has an aged, yellowed appearance, adding to the historical feel of the artwork.}\\

\textbf{(l)} \textit{An artistic representation of the planet Earth, with a swirl of musical notes in black ink encircling the globe. The Earth is depicted in vibrant blues and greens, indicating the oceans and continents, while the musical notes appear to dance around the planet's surface. The background of the drawing is a stark white, emphasizing the contrast and the harmony between music and the world.}\\

\textbf{(m)} \textit{In the midst of a vibrant garden, a cylindrical green cup stands alone on a stone path, its surface reflecting the bright afternoon sunlight. The cup, with a smooth finish, is surrounded by blossoming flowers and lush greenery. The shadows of nearby plants dance on the cup as gentle breezes sway their leaves.}\\

\textbf{(n)} \textit{An intricately detailed oil painting that captures the whimsical essence of a feline super math wizard. The cat, adorned with a wizard's hat and cape, is surrounded by floating mathematical symbols and equations. The rich textures of the brush strokes give depth to the cat's fur and the magical elements, creating a vivid and captivating scene.}\\

\textbf{(o)} \textit{Inside a warm room with a large window showcasing a picturesque winter landscape, three gleaming ruby red necklaces are elegantly laid out on the plush surface of a deep purple velvet jewelry box. The gentle glow from the overhead light accentuates the rich color and intricate design of the necklaces. Just beyond the glass pane, snowflakes can be seen gently falling to coat the ground outside in a blanket of white.}\\

\textbf{(p)} \textit{On a clear warm day, the sun radiates down on a sandy beach where a close-up of a wafer cone reveals chocolate ice cream beginning to melt down its textured sides. In the background, the sea glistens and reflects the sunlight, with gentle waves lapping at the shore. The ice cream's rich brown tones contrast sharply with the blue and turquoise hues of the ocean, creating a striking visual. Nearby, a few colorful beach umbrellas are dotted along the water's edge, offering shade to beachgoers.}\par

\endgroup

\textbf{Prompts for Figure~\ref{fig:bagel}.}
The prompts corresponding to columns (a)--(p) are listed below.\\

\begingroup
\small
\setlength{\parindent}{0pt}

\textbf{GenEval:}\\

\textbf{(a)} \textit{A photo of a potted plant and a boat.}\\

\textbf{(b)} \textit{A photo of a laptop left of a cow.}\\

\textbf{(c)} \textit{A photo of two teddy bears.}\\

\textbf{(d)} \textit{A photo of four vases.}\\

\textbf{(e)} \textit{A photo of a red bicycle.}\\

\textbf{(f)} \textit{A photo of a toaster.}\\

\textbf{(g)} \textit{A photo of a fire hydrant and a tennis racket.}\\

\textbf{(h)} \textit{A photo of a bowl and a skis.}\\

\medskip
\textbf{DPG-Bench:}\par
\textbf{(i)} \textit{A bright yellow tennis ball lies in stark contrast on the vibrant green of a freshly mowed grass court. The ball's fuzzy texture is highlighted by the sunlight, casting a small shadow on the neatly trimmed lawn. Nearby the baseline, the white chalk lines distinctly mark the boundaries of the playing field, creating a geometric harmony on the court.}\\

\textbf{(j)} \textit{An ornate royal carriage, painted in deep red with golden trim, stands prominently against a landscape blanketed in pristine snow. Behind it, the silhouettes of tall pine trees dusted with white can be discerned through the soft haze of a winter's day. In front of the carriage, the snow-covered ground glistens under the subtle light of the afternoon sun.}\\

\textbf{(k)} \textit{A rustic wooden table set outside, perhaps in a garden or patio area. On its surface, a pair of small birds are perched, casually observing their surroundings. The table shows signs of weathering, indicating it's been a part of the outdoor scenery for some time.}\\

\textbf{(l)} \textit{An anime-style illustration depicts a muscular, metallic tiger with sharp, angular features, standing on a rooftop. The tiger is in a dynamic pose, gripping a sleek, red electric guitar, and its mouth is open wide as if caught in the midst of a powerful roar or song. Above the tiger, a bright spotlight casts a dramatic beam of light, illuminating the scene and creating stark shadows on the surrounding rooftop features.}\\

\textbf{(m)} \textit{An intricately detailed oil painting captures the essence of a young badger, its fur rendered with fine brushstrokes that give it a tactile quality. The badger's snout is gently poised above a vibrant yellow rose, which stands out against a backdrop of muted green foliage. The contrast between the animal's coarse fur and the delicate petals of the rose is emphasized through the artist's skillful use of texture and color.}\\

\textbf{(n)} \textit{A surreal composite image showcasing the iconic Sydney Opera House with its distinctive white sail-like structures, positioned improbably beside the towering Eiffel Tower, its iron lattice work silhouetted against the night. The backdrop is a vibrant blue sky, pulsating with dynamic energy, where yellow stars burst forth in a dazzling display, and swirls of deeper blue spiral outward. The scene is bathed in an ethereal light that highlights the contrasting textures of the smooth, shell-like tiles of the Opera House and the intricate metalwork of the Eiffel Tower.}\\

\textbf{(o)} \textit{In a modern kitchen, a square, chrome toaster with a sleek finish sits prominently on the marble countertop, its size dwarfing the nearby red vintage rotary telephone, which is placed quaintly on a wooden dining table. The telephone's vibrant red hue contrasts with the neutral tones of the kitchen, and its cord coils gracefully beside it. The polished surfaces of both the toaster and the telephone catch the ambient light, adding a subtle shine to their respective textures.}\\

\textbf{(p)} \textit{
A cozy kitchen scene where a plush teddy bear is seated amongst a bunch of ripe bananas. The bananas are resting on a wooden countertop, which also features a variety of other groceries and kitchen utensils. In the background, a window allows natural light to illuminate the teddy bear's soft fur and the vibrant yellow of the bananas.}

\endgroup

\end{document}